\documentclass[acmtog]{acmart}

\AtBeginDocument{%
  \providecommand\BibTeX{{%
    \normalfont B\kern-0.5em{\scshape i\kern-0.25em b}\kern-0.8em\TeX}}}

\setcopyright{rightsretained}
\acmJournal{TOG}
\acmYear{2022}\acmVolume{41}\acmNumber{4}\acmArticle{120}\acmMonth{7}
\acmDOI{10.1145/3528223.3530079}

\acmSubmissionID{231}

\usepackage{url}

\begin{document}

\title{NIMBLE: A Non-rigid Hand Model with Bones and Muscles}


\author{Yuwei Li}
\affiliation{%
  \institution{School of Information Science and Technology, ShanghaiTech University}
  \city{Shanghai}
  \country{China}}
\additionalaffiliation{
  \institution{Shanghai Institute of Microsystem and Information Technology, Chinese Academy of Sciences; University of Chinese Academy of Sciences}
  \city{Shanghai}
  \country{China}
}
\email{liyw@shanghaitech.edu.cn}


\author{Longwen Zhang}
\affiliation{
  \institution{ShanghaiTech University, China and Deemos Technology}
  \city{Shanghai}
  \country{China}}
\email{zhanglw2@shanghaitech.edu.cn}
\email{zhanglw@deemos.com}

\author{Zesong Qiu}
\affiliation{%
  \institution{ShanghaiTech University}
  \city{Shanghai}
  \country{China}}
\email{qiuzs@shanghaitech.edu.cn}

\author{Yingwenqi Jiang}
\affiliation{%
  \institution{ShanghaiTech University}
  \city{Shanghai}
  \country{China}}
\email{jiangywq@shanghaitech.edu.cn}

\author{Nianyi Li}
\affiliation{%
\institution{Clemson University}
\city{Clemson}
\country{United States of America}
}
\email{nianyil@clemson.edu}

\author{Yuexin Ma}
\affiliation{%
\institution{ShanghaiTech University}
\city{Shanghai}
\country{China}}
\email{mayuexin@shanghaitech.edu.cn}

\author{Yuyao Zhang}
\affiliation{%
  \institution{ShanghaiTech University}
  \city{Shanghai}
  \country{China}}
\email{zhangyy8@shanghaitech.edu.cn}

\author{Lan Xu}
\affiliation{%
  \institution{ShanghaiTech University}
  \city{Shanghai}
  \country{China}}
\email{xulan1@shanghaitech.edu.cn}

\author{Jingyi Yu}
\authornote{corresponding author}
\affiliation{%
  \institution{ShanghaiTech University, China and DGene Digital Technology}
  \city{Shanghai}
  \country{China}}
\email{yujingyi@shanghaitech.edu.cn}

\renewcommand{\shortauthors}{Li, et al.}

\begin{abstract}

 Emerging Metaverse applications demand reliable, accurate, and photorealistic reproductions of human hands to perform sophisticated operations as if in the physical world. While real human hand represents one of the most intricate coordination between bones, muscle, tendon, and skin, state-of-the-art techniques unanimously focus on modeling only the skeleton of the hand. In this paper, we present NIMBLE, a novel parametric hand model that includes the missing key components, bringing 3D hand model to a new level of realism. We first annotate muscles, bones and skins on the recent Magnetic Resonance Imaging hand (MRI-Hand) dataset ~\cite{li2021piano} and then register a volumetric template hand onto individual poses and subjects within the dataset. NIMBLE consists of 20 bones as triangular meshes, 7 muscle groups as tetrahedral meshes, and a skin mesh. Via iterative shape registration and parameter learning, it further produces shape blend shapes, pose blend shapes, and a joint regressor. We demonstrate applying NIMBLE to modeling, rendering, and visual inference tasks. By enforcing the inner bones and muscles to match anatomic and kinematic rules, NIMBLE can animate 3D hands to new poses at unprecedented realism. To model the appearance of skin, we further construct a photometric HandStage to acquire high-quality textures and normal maps to model wrinkles and palm print. Finally, NIMBLE also benefits learning-based hand pose and shape estimation by either synthesizing rich data or acting directly as a differentiable layer in the inference network.
\end{abstract}



\begin{teaserfigure}
  \includegraphics[width=\textwidth]{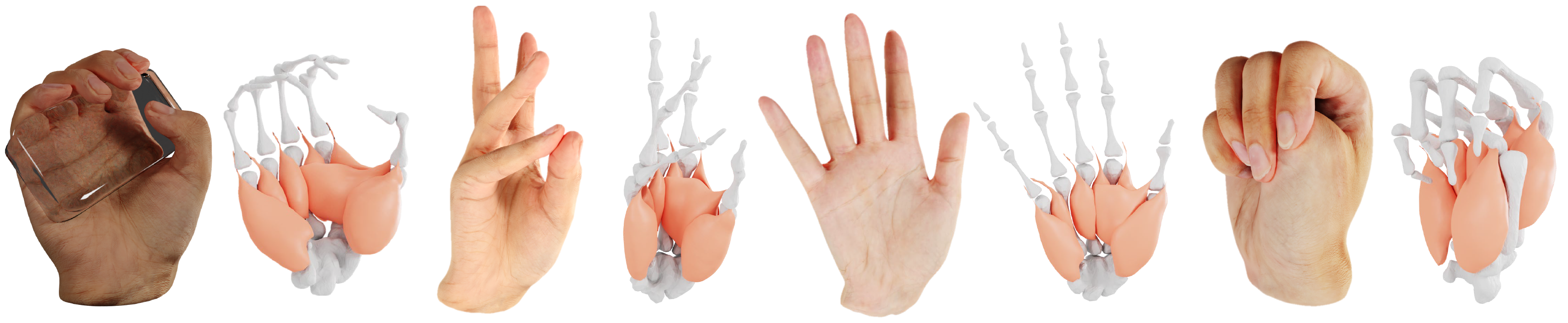}
  \caption{
  We present NIMBLE, a non-rigid parametric hand model that includes bones and muscles, bringing 3D hand model to a new level of realism. By enforcing the inner bones and muscles to match anatomic and kinematic rules, NIMBLE can animate 3D hands to new poses at unprecedented realism. }
  \label{fig:teaser}
\end{teaserfigure}

\maketitle

\section{Introduction}

In the production of animated feature films, VFX juggernauts spend most of their time and resources on rendering face, hair and skin complexion, as a shortcut or trick for creating realistic human and feelings. In contrast, renderings of other parts of human body, particularly human hands, are often glossed over. The reason is simple: VFX producers can easily direct the focus of audience attention away from hands and onto hair and skin, where they can generate the most life-like and thereby attention-grabbing visual effects. Indeed, we rarely see scenes featuring complex and dexterous hand movements. 
At the dawn of the Metaverse, however, emerging virtual reality consumer products will be decidedly more intimate, immersive and interactive and therefore demand life-like renderings on all body parts, especially hands. When users put on the head-mounted displays, their virtual hands should replace the physical ones to perform as many operations in the Metaverse as in real life. The dexterousness of human hands - the complex geometrical structures, the marvelous things and the subtle messages that fingers can construct, create and pass on when they move - define us humans as intelligent beings in both physical and digital worlds. 

Hands, however, are difficult to model. Through evolution, hand movements have become an intricate orchestration of bones, muscles, ligaments, nerves, and skins. Performing a specific gesture, for example, stems from the dragging and pulling of the hand muscles, which then drives the movements of the bones and eventually changes the shape and appearance caused by muscle deformation and skin wrinkling. To faithfully reproduce 3D hand movements, it is critical to not only model each individual component but more importantly to model their delicate coordination. In contrast to tremendous efforts on 3D human body modeling  \cite{loper2015smpl,hirshberg2012coregistration,anguelov2005scape,pons2015dyna}, research on photo-realistic 3D human hands is rather limited. By far, the majority of prior art has focused on modeling the skeleton of the hand. In fact, 
the most adopted hand model, MANO \cite{romero2017embodied}, defines skeleton in terms of empirical joint locations without taking into account anatomical bone structure.
The recent PIANO \cite{li2021piano} model extends MANO by employing anatomically correct bone structures and shapes, as well as joints to connect the bones. PIANO manages to produce more convincing movements but the resulting appearance still lacks realism as it ignores muscles and skins that deform along with the bones. In reality, even a routine posture such as grabbing or holding a fist requires 24 muscle groups surrounding the hand and the wrist \cite{handanatomy} working together to execute extension and flexion. Shape deformations of these muscle groups subsequently affect the appearance of their covering skins and the overall realism in virtual hand rendering. By far, the graphics community still lacks a reliable and practical musculoskeletal parametric hand model. 

In this paper, we present NIMBLE, a Non-rIgid hand Model with skins, Bones, and muscLEs, to bring 3D modeling of dynamic hands to a new level of realism. Our work is enabled by the recent Magnetic Resonance Imaging hand (MRI-Hand) dataset ~\cite{li2021piano} that captures 35 subjects (19 male 16 female), with 50 different hand poses, along with annotated segmentation labels of bones. We first conduct comprehensive segmentation annotations to further identify muscles and skins on the MRI-Hand dataset. Similar to how SMPL \cite{loper2015smpl} starts with an T-pose for modeling comprehensive movements, 
NIMBLE utilizes a general rest hand pose as the template that includes 20 bones, 7 muscle groups and a skin mesh. In particular, for the sake of computation and rendering efficiency, we cluster 24 anatomic muscles groups to 7 while preserving as much physical meaning as possible.  

To derive a new parametric hand model, NIMBLE uses triangle meshes to model bones and tetrahedral meshes for deformable muscles and skins. The formulation manages to model shape deformations while maintaining structural rigidity. To register the internal structures under the rest pose to the ones in MRI-Hand, we present a multi-stage registration technique that leverages the benefits of pose initialization, the interdependence of hand structures, and physical-based simulations. Our technique accurately models the elasticity of deformations without sacrificing pose accuracy. The registered results lead to a new parametric model containing bones, muscles, skin template meshes, kinematic tree, shape blend shapes, pose blend shapes, and a joint regressor. In addition to parameter learning, we further impose penalty terms to avoid collision and enforce physically correct muscle deforms. 

Applications of NIMBLE span from geometric modeling and rendering to visual inference. We first demonstrate applying NIMBLE to animate 3D hands of different shapes and under arbitrary poses. NIMBLE provides an unprecedented level of realism by enforcing inner bones and muscles to match anatomic and kinematic rules. To further enhance visual realism, we construct a photometric appearance capture system called HandStage, analogous to the USC LightStage~\cite{debevec2012light}, to acquire high-quality textures with appearance details (e.g., normal maps), including wrinkles and palm print. These appearance details further improve visual fidelity and adapt to new lighting conditions. NIMBLE, as a parametric model, can be further integrated into state-of-the-art animation systems and rendering engines such as Blender and Unity for VR and AR applications. 
Furthermore, parameter-represented NIMBLE benefits learning-based hand pose and shape estimation on both MRI and RGB imagery, by either synthesizing rich data with varying shapes and poses or acting directly as a differentiable layer in the inference network.

To summarize, our main contributions include:
\begin{itemize} 
	\setlength\itemsep{0em}
	\item We exploit the recent MRI-Hand dataset to conduct the complete segmentation annotations for bones, muscles, and skins, as well as auto-registered meshes by optimization with physical constraints.
	
	\item We derive a parametric NIMBLE model by iterating between shape registration and parameter learning, using new penalty terms to guarantee physically correct bone movements and muscle deformations.
	
	\item We demonstrate using NIMBLE for anatomically correct digital hand synthesis, motion animation and photorealistic rendering. The results benefit many downstream tasks including pose and shape inference, visual tracking, etc.

	\item We make available our NIMBLE model and annotation data at \url{https://reyuwei.github.io/proj/nimble}.
\end{itemize} 

\section{Related Work}
In this section, we survey closely related works and discuss the relationship with the proposed work.

\paragraph{Parametric Models.} With parametric modeling, low-dimensional parametric space is estimated to approach human body geometry.
Many existing methods incorporate linear blend skinning, which deforms a mesh based non-linear combination of rigid transformations of associated bones, on top of various skeletal representations. With skinning weights carefully designed, these presentations can produce reasonable deformations for articulated body tissues. A pioneer work on 3D morphable face is proposed by~\cite{morphableMD}. Since then numerous methods have learned 3D face shape and expression from scanning data~\cite{LearnID_Pose,sota_3dface}. The advantage of such geometric models is their ability to represent variety of face shapes and wide range of expressions. Unlike most face models that focus only on the facial region, recent popular models FLAME~\cite{flame} and its extension DECA~\cite{feng2021learning} consider the whole head and neck regions instead. With the entire head, the authors were able to assume a simulated jaw joint to achieve large deformed facial pose and expressions. The availability of 3D body scanners enabled learning of body shape from scans. Since the CAESAR dataset opened up the learning of body shape~\cite{reconst_pm}, most early works focus on modelling only body shape (varying with identity) using subjects scanned in roughly the same pose. Combining body shapes from a group of population and different poses of a single subject,~\cite{anguelov2005scape}  learned a factored model of both body shape and pose based on triangle deformations. Following this work, many human body parametric models were built using either triangle deformations~\cite{pons2015dyna,hirshberg2012coregistration} or vertex-based displacements~\cite{10.1145/1730804.1730809,loper2015smpl}, however all these works focus on modeling body shape and pose without the hands or face. Comparing with face and body, human hands are more complex for parametric modelling due to the extreme flexibility of hand motion. Early 3D hand models are typically not learned but based on shape primitives~\cite{10.5555/2532129.2532141,Oikonomidis2011EfficientM3,Schmidt2014DARTDA},  reconstructed with multiview stereo with fixed shape~\cite{10.1007/978-3-642-33783-3_46,10.1007/s11263-016-0895-4}, and use non-learned per-part scaling parameters~\cite{modelbased3dhand}, or use simple shape spaces~\cite{10.1145/2980179.2980226}. Only recently~\cite{romero2017embodied,Khamis2015LearningAE,li2021piano} proposed learned hand models.~\cite{Khamis2015LearningAE} collect partial depth maps of 50 people to learn a model of shape variation, however they do not capture a pose space.~\cite{romero2017embodied} on the other side learn a parametric hand model (MANO) with both a rich shape and pose space using 3D scans of 31 subjects in up to 51 poses, following the SMPL~\cite{loper2015smpl} formulation.~\cite{li2021piano} built up a parametric hand bone model from MRI data, which drove the hand shape and pose using real bone and joint structures.

\paragraph{Hand Models.} Hand modeling is an essential topic in computer graphics. Many hand models have been proposed and are summarized in Table \ref{tab:model_compare} categorizing according to their intuitive emphasis of the hand inner biomechanical structures~\cite{albrecht2003construction,wang2019hand,wang2020modeling,li2021piano} or the hand outer shape, color and texture~\cite{romero2017embodied,DeepHandMesh,qian2020html}. 

\paragraph{Hand Inner Biomechanical Model.} Hand structure and functions are biomechanically complex. Therefore, physical-based kinetic simulation is essential to model hand pose and shapes. 
For example, in early works, using simulated underlying hand skeleton to generate a solid hand surface mesh constrain~\cite{10.1145/1073368.1073412,10.1145/2019627.2019640,10.1145/2508363.2508427}. Not just bones, hand skin and tendons are also widely considered to refine the visual appearance or control of hand articulation~\cite{10.1145/1360612.1360682,10.1145/2461912.2462008,10.1145/2766987}. 
Musculotendon modeling and simulation~\cite{kadlevcek2016reconstructing,lee2018dexterous,abdrashitov2021interactive} have also been studied on human body.
More recently, for achieving real-human-like hand animation performance, researchers paid more attention on anatomical structures instead of simulated models.~\cite{mirakhorlo2018musculoskeletal} comprehensively modeled biomechanical hand model based on detailed measurements from a hand specimen, yet it is not differentiable and can not be embedded in deep learning frameworks.~\cite{wrist_pmodel_16} proposed a statistical wrist shape and bone model for automatic carpal bone segmentation.
~\cite{wang2019hand} acquired a single-subject complete hand bone and muscle~\cite{wang2020modeling} anatomy in multiple poses using magnetic resonance imaging (MRI), to build an anatomy-correct hand bone rig of a target performer. However, it suffers from time-consuming and user-specified bone segmentation operations, which is impractical to apply to various individuals for building a parametric model.~\cite{li2021piano} construct a parametric hand bone model named PIANO from multiple-subject and multiple-pose MRI acquisitions, which is physically precise and differentiable. It can be applied in deep neural networks for computer vision tasks. However, it still lacks more comprehensive anatomical structures such as muscle to support more realistic hand outer surface generation.

\paragraph{Hand Outer Appearance Model.} Fully driven by the underlying biomechanical hand structures, skinning technique still acts as indispensable procedure in generating high-fidelity hand animation. 
~\cite{10.1145/344779.344862} proposed Pose-Space Deformation (PSD) that combines skeleton subspace deformation~\cite{10.5555/102313.102317} with artist-corrected pose shapes. It is widely used in industry due to its speed, simplicity and the ability to incorporate real-world scans and arbitrary artist corrections.
Kry and his collaborators~\shortcite{kry2002eigenskin} further proposed to use Principal Component Analysis (PCA) to represent the large set of pose corrections.
Recently,~\cite{romero2017embodied} augmented an LBS-based hand surface model with statistical individual- and pose-dependent parametric correctives, constructing a system referred as MANO~\cite{romero2017embodied}. MANO has been widely used in variety of hand fitting and tracking scenarios including hand interactions~\cite{hasson2019learning,mueller_siggraph2019} and single RGB image hand pose estimation~\cite{DBLP:journals/corr/abs-1904-04196}. These approaches are fully constrained by the underlying MANO model, which lacks real biomechanically correct constrains from the underlying hand tissue, and thus may fail to replicate subtle details of hand geometry like creases and bulging. Inspired by the interesting works, we construct the first complete parametric hand model with bone, muscle, and skin.
By involving inherent kinematic structures and considering physically precise constraints, NIMBLE enables authentic hand shape and appearance generation and differentiable training for many down stream tasks.

\begin{table*}
    \centering
    \caption{NIMBLE vs. existing hand models.}
    \begin{tabular}{c|c|c|c|c|c|c|c}
    \hline
        Model  & Parametric Model & Skin & Bone & Muscle & Shape & Pose & Appearance\\ \hline \hline
        \cite{albrecht2003construction} & \textcolor{red}{$\times$} & \textcolor{green}{\checkmark} & \textcolor{green}{\checkmark} & \textcolor{green}{\checkmark} & \textcolor{red}{$\times$} & \textcolor{green}{\checkmark} & \textcolor{red}{$\times$} \\ \hline
        \cite{wang2019hand} & \textcolor{red}{$\times$} & \textcolor{green}{\checkmark} & \textcolor{green}{\checkmark} & \textcolor{red}{$\times$} & \textcolor{green}{\checkmark} & \textcolor{green}{\checkmark} & \textcolor{green}{\checkmark} \\ \hline
        \cite{wang2020modeling} &\textcolor{red}{$\times$} & \textcolor{red}{$\times$} & \textcolor{red}{$\times$} & \textcolor{green}{\checkmark} & \textcolor{green}{\checkmark} & \textcolor{red}{$\times$} & \textcolor{red}{$\times$} \\ \hline
        MANO~\cite{romero2017embodied} & \textcolor{green}{\checkmark} &\textcolor{green}{\checkmark}  & \textcolor{red}{$\times$} & \textcolor{red}{$\times$} &\textcolor{green}{\checkmark} &\textcolor{green}{\checkmark} & \textcolor{red}{$\times$} \\ \hline
        HTML~\cite{qian2020html} & \textcolor{green}{\checkmark} &\textcolor{green}{\checkmark}  & \textcolor{red}{$\times$} & \textcolor{red}{$\times$} &\textcolor{green}{\checkmark} &\textcolor{green}{\checkmark} & \textcolor{green}{\checkmark} \\ \hline
        PIANO~\cite{li2021piano} & \textcolor{green}{\checkmark} & \textcolor{red}{$\times$} &\textcolor{green}{\checkmark} & \textcolor{red}{$\times$} &\textcolor{green}{\checkmark} &\textcolor{green}{\checkmark} & \textcolor{red}{$\times$} \\ \hline
        NIMBLE (Ours) & \textcolor{green}{\checkmark} & \textcolor{green}{\checkmark} &\textcolor{green}{\checkmark} &\textcolor{green}{\checkmark} &\textcolor{green}{\checkmark} &\textcolor{green}{\checkmark} & \textcolor{green}{\checkmark} \\ \hline
    \end{tabular}
    \label{tab:model_compare}
\end{table*}

\section{Overview}

\begin{figure}
    \centering
    \includegraphics[width=\linewidth]{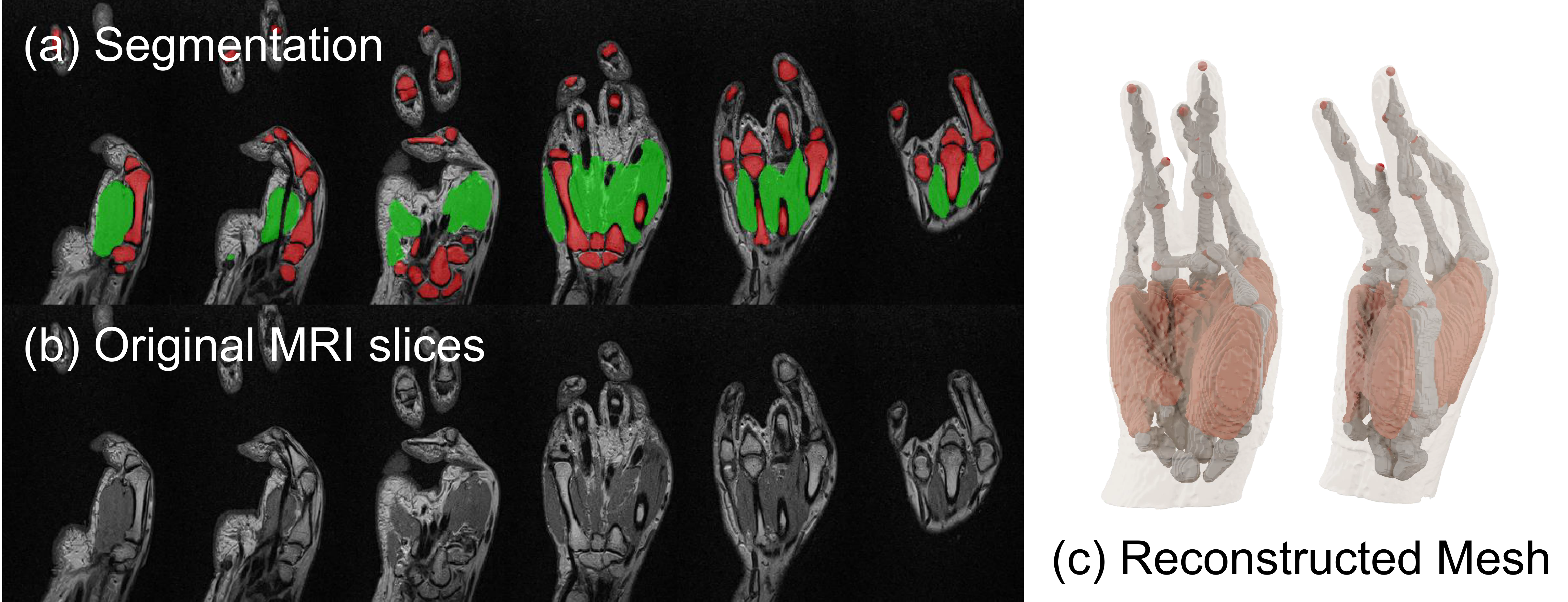}
    \caption{(a) MRI annotation of bone and muscle mask. (b) Original slices. (c) Reconstructed bone, muscle and skin mesh, joints are visualized in red.}
    \label{fig:mri_annot}
\end{figure}

We present a novel method for \textit{Non-rIgid hand paraMetric modeling on Bone and MuscLE (NIMBLE)}. To our best knowledge, this is the first parametric hand anatomy-based algorithm that can simultaneously model the interior hand kinematic structure and the exterior hand shape with the high-fidelity appearance of individuals. 

\begin{figure*}[h]
    \centering
    \includegraphics[width=\linewidth]{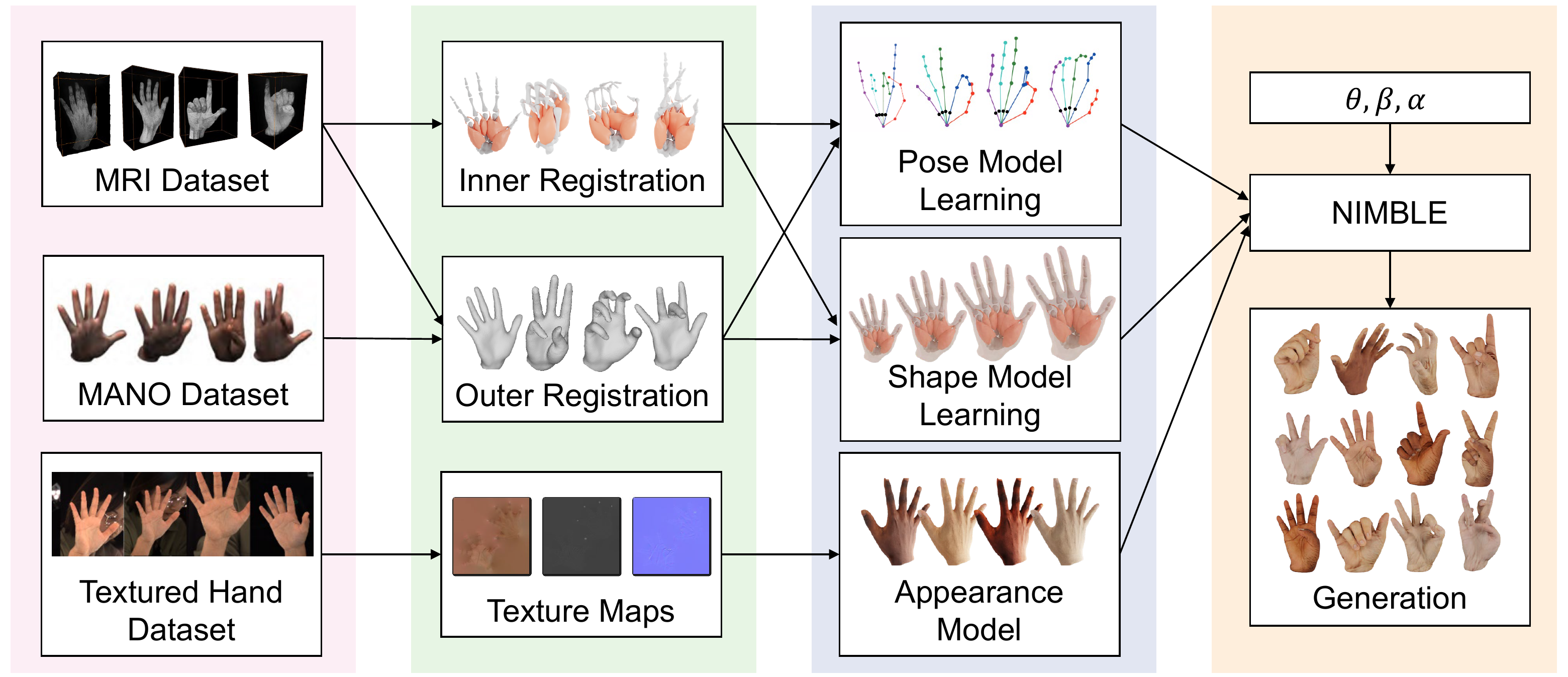}
    \caption{Overview of building NIMBLE, which includes inner and outer registration and parametric model learning. As well as NIMBLE application for synthetic hand generation and photorealistic rendering. $\theta$, $\beta$, $\alpha$ are parameters that control model pose, shape and appearance.}
    \label{fig:whole_pipeline}
\end{figure*}

NIMBLE is developed on a large amount of hand data with annotated inner and outer structural features. Specifically, we use the MRI hand dataset from \cite{li2021piano} and further annotate the MRI data to segment out the muscle and skins from the original annotations, as shown in Figure \ref{fig:mri_annot}. The hand appearances are represented as textures with diffuse, normal, and specular maps, collected by a photometric appearance capture system called HandStage.
We then build a parametric hand model by registering the hand template to all the interior and exterior hand features and photometric appearances in the dataset. 
After registration, we extend the general hand modeling pipeline followed by \cite{romero2017embodied,li2021piano}, so that \textit{NIMBLE} can learn a complete anatomy-based hand model with bone, muscles and textures by iteratively fitting the hand template to the multi-modal data and regenerating the parametric model from the registered multi-model features. 
The pipeline of our method is shown in Figure \ref{fig:whole_pipeline}.
The rest of the paper is organized as follows: we firstly introduce our data collection and annotation in Section \ref{sec:data_collect}. Next, we show our model formulation in Section \ref{sec:model_formulation}, followed by a physically based registration in Section \ref{sec:registration} and multi-stage parameter learning on shape and pose in Section \ref{sec:pm_train}. After having the hand template, we attach hand appearance to get photo-realistic rendering effect, as discussed in Section \ref{sec:rendering}. In Section \ref{sec:experiment}, we evaluate the effectiveness of NIMBLE by numerous 3D hands of different poses, shapes, appearances, and photorealistic rendering conditions. We also show that our method can be easily fitted into the hand inference pipeline from various input.

\subsection{MRI Data Collection and Preparation}
\label{sec:data_collect}

The dataset from \cite{li2021piano} contains 200 hand MRI volumes spanning 50 different hand postures of 35 individual subjects. However, it only provides annotation on the bone mask and 25 joint positions per hand. In this paper, we regenerate a fine-grained bone segmentation mask on each MRI volume using radial basis functions and joint annotation. 
Additionally, we handcraft binary muscle masks on each volume slice using Amira\cite{webamira}. For time efficiency, we only annotate large and notable muscle areas on each MRI scan, and our model registration algorithm, as discussed  in Section \ref{sec:registration}, can automatically fill in the missing parts.
As for skin annotation, we use an automatic thresholding method \cite{Otsu1979ATS} to extract skin mask. 
Next, we extract the iso-surface of bone, muscle and skin by applying the Marching Cubes algorithm \cite{lorensen1987marching} to get rough hand meshes, as shown in Fig \ref{fig:mri_annot}(c).
\section{NIMBLE}
\label{sec:data_capture}

Modeling photo-realistic hand is challenging as the interior structure of the hand, including muscles and bones, are unknown, whereas the non-rigid and elastic property of muscles and the biomedical connection between muscles and bones can largely determine the exterior shape of hands. State-of-the-art solutions have neglected this connection and thus failed at capturing realistic movement like muscle and skin bulging.
We address this issue by jointly modelling the internal hand structure with outer appearances while considering the physical constraints of the relative motion between muscles, bones and skins.

\subsection{Model Formulation}
\label{sec:model_formulation}

\begin{figure}
    \centering
    \includegraphics[width=0.9\linewidth]{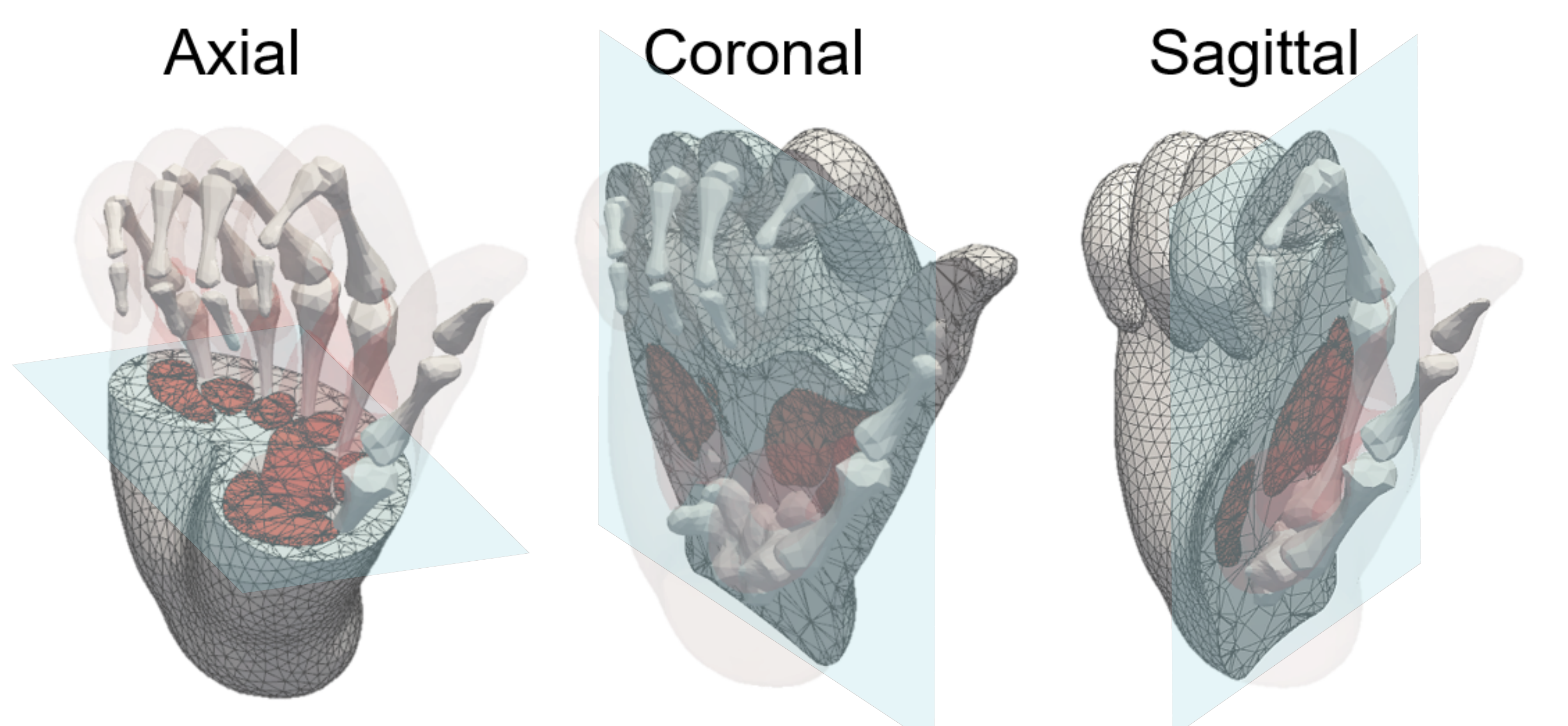}
    \caption{Three cutaway views of the tetrahedral mesh of our template hand.}
    \label{fig:model_template}
\end{figure}

\begin{table}[]
    \caption{NIMBLE template mesh details. Number of semantic parts, vertices, mesh faces and tetrahedrons. }
    \begin{tabular}{c|c|c|c|c}
    \hline
    Tissue    & \# parts  & \# vertices & \# faces  & \# tetrahedron    \\ \hline \hline
    Bone      &   20       &  3345      & 6610      &   -            \\ \hline
    Muscle    &   7        &  5635      & 10512     & 15986          \\ \hline
    Skin      &   1        &  5990      & 9984      & 19562          \\ \hline
    \end{tabular}
    \label{tab:template_v}
\end{table}

The general formulation of NIMBLE is defined as follows:
\begin{equation}
    \mathcal{N}({\theta}, {\beta}, \alpha) = \{\mathcal{G}(\theta, \beta),  \mathcal{A}(\alpha)\},
    \label{eqn:full_formulation}
\end{equation}
where $\mathcal{G}$ denotes the hand geometry, and $\mathcal{A}$ models the hand appearance. $\theta$, $\beta$, $\alpha$ are parameters controlling hand pose, shape and appearance, respectively. In this section, we will focus on the model of hand geometry.

To generate an accurate hand template, we extend the PIANO \cite{li2021piano} pipeline, which only considers the bone structures, shapes and joints, and add the muscle and skin features in the template formulation $\mathcal{G}$: 
\begin{equation}
    \mathcal{G}({\theta}, {\beta}) = LBS(\mathcal{W}, J_p({\beta}), {\theta}, \mathbf{T}_p({\theta}, {\beta})).
    \label{eqn:formulation}
\end{equation}
where $LBS(\cdot)$ demotes the Linear Blend Skinning (LBS) function; $\mathcal{W}$ is the learned skinning weight of $LBS(\cdot)$; $J_p$ represent the joint locations; ${\theta}$ is an array of joints rotation axes and angles; ${\beta}$ is the PCA coefficient vector of the shape space; and $\mathbf{T}_p$ is a person-specific hand template mesh.
In another word, we can formulate $\mathcal{G}$ of arbitrary individuals by Eqn.~\ref{eqn:formulation}, as long as we know $(\mathcal{W},J_p,\beta,\theta,\mathbf{T}_p)$. 
Specifically, $J_p$ is defined by a joint regressor $\mathcal{J}$ that maps the bone mesh vertices $\overline{\mathbf{T}}_b$ to joint locations by taking into account the shape parameters $\beta$, we refer readers to \cite{li2021piano,loper2015smpl} for details. Note that we use bone mesh because joints are the essential rotation center of bone segments, which is invariant to skin and muscle shape.

The personalized template $\mathbf{T}_p$ is a linear combination of general hand template $\overline{\mathbf{T}}$, pose blend shape $B_P$ and shape blend shape $B_S$ (Eqn. 2,3,and 4 of \cite{romero2017embodied}), where $B_P$ is the multiplication of pose blend shapes ${\mathcal{P}}$ and pose rotation matrix, and $B_S$ is the multiplication of orthonormal PCA of shape blend shape $\mathcal{S}$ and $\beta$. $B_P$ and $B_S$ can correct artifacts introduced by $LBS(\cdot)$ by adding vertex offsets to the general template $\overline{\mathbf{T}}$. We use the same number of rotation joint $K=19$ as in PIANO \cite{li2021piano}. In the following paper, we use $\{\mathcal{W},\mathcal{J},\beta,\theta,\overline{\mathbf{T}},\mathcal{P},\mathcal{S}\}$ to parameterize $\mathcal{G}$.

Unlike the popular surface modeling methods \textit{i.e.}, SMPL \cite{loper2015smpl} and MANO \cite{romero2017embodied}, we define the general hand template $ \overline{\mathbf{T}}$ by jointly considering the bones, muscles and skin mesh as a whole set $ \overline{\mathbf{T}} = \{ \overline{\mathbf{T}}_b,  \overline{\mathbf{T}}_m,  \overline{\mathbf{T}}_s\}$. As for bones, we adopt the triangular mesh settings in PIANO \cite{li2021piano} due to the rigid deformation property of bones. 
We use tetrahedral mesh to model muscle and skin so that NIMBLE can capture the non-rigid motion effects such as muscle bulging and wrinkled skin. 
Given that each hand has about 24 major muscles, which will significantly increase the computation and rendering cost if NIMBLE models all the muscles, we follow \cite{schwarz1955anatomy,erolin2016does} and anatomically integrate the muscle groups into 7 ones based on their functional and physical properties. 
As for skin template $\overline{\mathbf{T}}_s$, we use MANO topology as an initialization. 
We manually register bone, muscle and skin mesh to the same rest pose in the MRI dataset using \cite{webr3ds}, so that the three components are into the same physical space. 
The process takes less than 10 minutes and is only performed once.
We then remesh the registered triangular mesh of seven muscles and skin using isotropic explicit remeshing algorithm \cite{alliez2003isotropic} with a target edge length of 3mm, and create tetrahedral meshes using Tetgen \cite{si2015tetgen}.
The model details are listed in Table \ref{tab:template_v}, and Figure \ref{fig:model_template} shows three cutaway views of our template mesh with tetrahedrons. Our complete hand template mesh consists of $\overline{\mathbf{T}}$ with 14970 vertices and 27106 faces.
%

\subsection{Muscle Registration}
\label{sec:registration}

\begin{figure*}
    \centering
    \includegraphics[width=\linewidth]{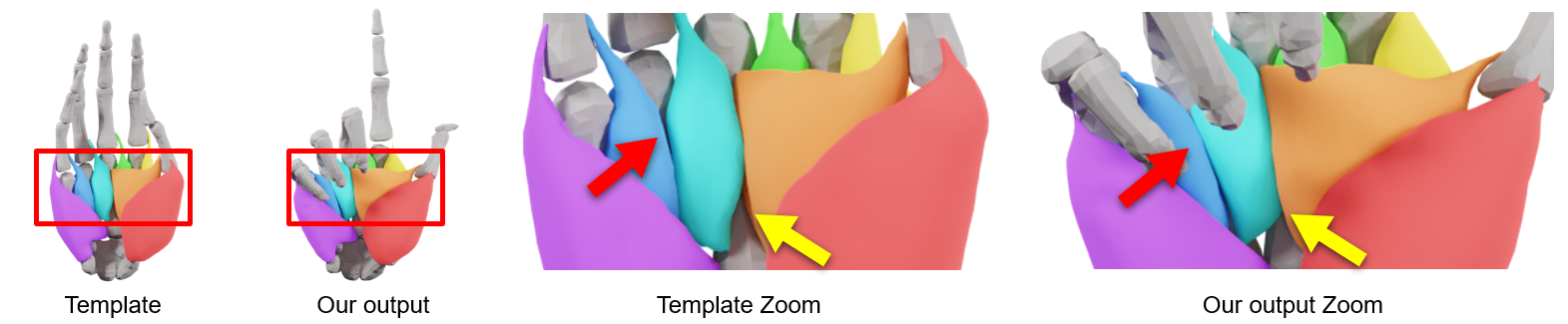}
    \caption{Registered hand muscles from MRI segmentation. Observe that the muscle around the arrows become thicker and tighter after registration.}
    \label{fig:reg_results}
\end{figure*}

Before training the parametric hand model, we need to first register the general template mesh $\overline{\mathbf{T}}$ 
to scale the dataset within the same topology. However, mesh registration is an open question of long-standing, let alone our goal is to register meshes of low resolution from a large MRI dataset. 
In this paper, we bypass the manual landmark labelling method \cite{wang2020modeling}, and propose a physically based multi-stage registration algorithm that can model accurate poses with high-quality elastic non-rigid deformation. We adopt similar registration pipelines for muscle, bone, and skin. Here, we present muscle registration in detail, and briefly discuss its difference compared with bone and skin registrations. 

Generally, our registration pipeline consists of two steps: pose initialization and iterative refinement. 
Pose initialization is to provide a good initial alignment to account for the highly nonlinear deformation of muscle.
After initialization, we iteratively update the mesh vertex offset so that the deformed template {$\hat{\mathbf{T}}$}
best matches the target one from MRI scans.
\\

\paragraph{Pose Initialization.}
We use a simplified parametric model to initialize pose parameter $\tilde{\theta}$:
\begin{equation}
    \mathcal{G}({\tilde{\theta}}) = LBS(\tilde{\mathcal{W}},\tilde{J},{\tilde{\theta}},\tilde{\mathbf{T}}).
\end{equation}
where we removed all the shape relevant parameters in Equation (\ref{eqn:formulation}), including the shape parameter $\beta$, the shape blend shape $\mathcal{S}$, and the pose blend shape $\mathcal{P}$. We use $\{\tilde{J},\tilde{\mathbf{T}}\}$ to indicate the trimmed $\{J_p,\mathbf{T}_p\}$. $\tilde{\mathcal{W}}$ is the skinning weight of this LBS function, which has been initialized by radial basis functions (RBF) according to template joint positions \cite{rhee2007soft}. 
{We minimize the L2 joint error between the posed template and target joint annotation and solve for the inverse kinematics to obtain the initial pose $\tilde{\theta}$.}
\\

\paragraph{Iterative Refinement.}
Then, we perform non-rigid registration to align the hand model at a finer scale.
We formulate this as an energy minimization problem to match deformed template muscle mesh $\hat{\mathbf{T}}$ to target muscle mesh $M$.
The objective function for non-rigid registration is defined as:
\begin{equation}
    E(\hat{\mathbf{T}};M) = E_{geo} + E_{reg} + E_{ne} + E_{att} + E_{icol} + E_{ecol}.
    \label{eqn:reg_full}
\end{equation}
where $E_{geo}$ is geometry term, $E_{reg}$ is regularization term, $E_{ne}$ is non-rigid elasticity term, $E_{att}$ is the attachment constraints and $E_{icol}$, $E_{ecol}$ are the internal/external collision penalties. We will discuss each term and their benefit in details.


\noindent\textbf{Geometry Term $E_{geo}$}
Inspired by the surface tracking algorithms \cite{xu2019flyfusion,newcombe2015dynamicfusion,smith2020constraining}, we use vertex distance and normal angle error to measure the distance between template mesh $\hat{\mathbf{T}}$ and target mesh $M$:
\begin{equation}
    E_{geo} = \lambda_d \delta_d(\hat{\mathbf{T}}, M) + \lambda_{dn}\delta_{dn}(\hat{\mathbf{T}}, M),
    \label{eqn:geo}
\end{equation}
where $\delta_d(\cdot)$ measures the Chamfer Distance~\cite{chamfer} between two meshes and $\delta_{dn}(\cdot)$ computes the angle between the corresponding vertex normal.
The first term pulls the template vertex to match with the nearest target vertex, while the second term adds a normal penalty to prevent the template from fitted to the opposite vertex normal.


\noindent\textbf{Regularization Term $E_{reg}$}
The regularization term consists of three components, \textit{i.e.}, rigidity regularizer $E_{rig}$, face normal consistency regularizer $E_{fn}$, and edge length regularizer $E_{edge}$:  
\begin{equation}
    E_{reg} = \lambda_a E_{rig} + \lambda_{fn} E_{fn} + \lambda_e E_{edge}.
    \label{eqn:reg}
\end{equation}

$E_{rig}$ regularizes the deformation of vertices in $\hat{\mathbf{T}}$ by comparing the deformation degree of adjacent vertices to avoid implausible shapes in unobserved regions. Instead of directly regulating the mesh node's rotation, we add constraints on vertices:
\begin{equation}
    E_{rig} = \sum_{v_i\in \hat{\mathbf{T}}} \sum_{v_j \in \hat{\mathbf{T}}} w_{ij}(\mathbf{D}_i v_j - \mathbf{D}_j v_j),
    \label{eqn:smooth}
\end{equation}
where $\mathbf{D}_i$ represents the deformation of vertex $v_i$, and $w_{ij}$ is the weight between vertex $v_i$ and $v_j$: $w_{ij} = e^{-\|v_i - v_j\|_2^2 / 2\sigma^2}$. Higher $w_{ij}$ corresponds to closer distance, and thus higher impact.

To regularize the moving direction of vertices, we adopt the face normal consistency term $E_{fn}$ and edge length term $E_{edge}$ from \cite{wang2018pixel2mesh} to further ensure mesh surface smoothness and avoid flying vertices.
Specifically, the $E_{fn}$ computes the angle between the normal of each pair of neighbouring faces to ensure $\hat{\mathbf{T}}$ of consistent face normal and smooth surface.
$E_{edge}$ penalizes flying vertices that cause long edges by minimizing the average edge length.


\noindent\textbf{Non-rigid Elasticity Term $E_{ne}$}
To capture non-rigid deformation of hands,
we define $E_{ne}$ using the Neo-Hookean elastic function, which has been proven effective for muscle and flesh simulation in \cite{smith2020constraining,smith2018stable}:
\begin{equation}
    E_{ne} = \lambda_{ne} \sum_i V_i \psi(F)
\end{equation}
where $V_i$ denotes the $i^{th}$ tetrahedron's volume and $\psi(F)$ can be viewed as energy density. $\psi(\cdot)$ ensures the deformation gradient $F$ to be identical and thus can effectively prevent large changes and heavy self-collisions of muscles. Please refer to \cite{smith2020constraining} for the complete formulation of  $\psi(F)$.


\noindent\textbf{Attachment Constraints $E_{att}$}
To ensure that muscles are attached to their corresponding bones properly, we use $E_{att}$ to guarantee corresponding attachment points on the mesh surface:
\begin{equation}
    E_{att} = \lambda_{att} || \mathbf{A}_m - \mathbf{A}_b ||_2^2, 
\end{equation}
where $\mathbf{A}_m$ and $\mathbf{A}_b$ are hand-crafted attachment points matching on muscle and bone mesh.


\noindent\textbf{Internal Collision $E_{icol}$}
To avoid hand mesh self-penetration, {similar to \cite{hirota2001implicit}}, 
we penalize the internal collision by:
\begin{equation}
    E_{icol} = \lambda_{col} \sum_{v_{in}} ||n_{tar}^T \cdot (v_{in} - v_{tar})||^2_2.
    \label{eqn:icol}
\end{equation}
where $v_{in}$ refers to the interior penetration vertex, $v_{tar}$ is the target surface position, and $n_{tar}$ is the corresponding surface normal $n_{tar}$.
Due to the large search space of $v_{in}$ and $v_{tar}$, $E_{icol}$ can only handle small collisions. We therefore add an additional normal and distance filter to shrink the search space. 
Specifically, we discard collisions with normal angle larger than $90^{\circ}$ in $E_{icol}$ to remove large self-collision and finger penetrating the palm in $\hat{\mathbf{T}}$.

\noindent\textbf{External Collision $E_{ecol}$}
External collision happens between muscle to muscle and  muscle to bone. To eliminate this, we use the contact loss proposed in \cite{hasson2019learning}:
\begin{equation}
    E_{ecol} =  \lambda_{rep} E_{rep} + (1-\lambda_{rep}) E_{attr},
    \label{eqn:ecol}
\end{equation}
where $E_{rep}$ is a repulsion term, that measures the point-to-plane distance, and
$E_{attr}$ is a attraction term computing the point-to-point distance of correspondence vertices.  
$E_{rep}$ detects interpenetration points and pushes them towards the target mesh surface, the attraction term finds close vertices and force them to come into contact. 
By doing so, $E_{ecol}$ forces the muscle groups and bones to be adjacent without colliding each other.
\\

\paragraph{Bone and Skin Registration.}
The pipeline of bone and skin registration are similar to muscle's,
except that we use different term combination and balancing weights. 
For bone registration, we omit the non-rigid elasticity term $E_{ne}$ in Equation (\ref{eqn:reg_full}), considering the rigid deformation property of bones.
For skin registration, which also requires non-rigid deformation constrains, we use all the terms in Equation (\ref{eqn:reg_full}). The core difference is that we use larger weights on geometry term $E_{geo}$ to align skins, as the skin annotations are more reliable than muscles in the MRI dataset.

\subsection{Parameter Learning}
\label{sec:pm_train}
After registration, we have an initialized model $\{\tilde{\mathcal{W}},\tilde{\mathcal{J}},{\tilde{\theta}},\tilde{\mathbf{T}}\}$. The general template mesh $\overline{T}$ and hand scans in the MRI dataset have been aligned to the same topology.
Consequently, for each subject $S_i$ of hand pose $P_j$, we can generate a aligned mesh $g_{ij}$.
We then set out to train $\mathcal{G}=\{\mathcal{W},\mathcal{J},\beta, \theta,\overline{\mathbf{T}},\mathcal{P},\mathcal{S}\}$. 
Note that bones, muscles, and skins follow different anatomical and physical properties during shape and pose change. Therefore, given MRI scans, we train $\mathcal{G}$ by a multi-stage strategy to disentangle deformations by pose and shape. Nevertheless, the public MRI datasets only contain limited numbers of hand poses due to the high cost and time-intensity of MRI data acquisition. Thus, we further optimize $\mathcal{G}$ using additional hand scans from large image-based dataset to extend our pose variance.
\\

\paragraph{Learning on MRI dataset.}
Given MRI scans, we train $\mathcal{G}$ through three stages, \textit{i.e.,} the pose stage, the shape stage, and the parameter stage. For each stage, we only update certain parameters while keeping the rest fixed. The objective function is defined as follows:
\begin{equation}
    \begin{aligned}
        E_1 =~ & E_{po}(\theta|\mathcal{G}) + E_{sh}(\mathcal{J},\beta,\overline{\mathbf{T}},\mathcal{S}|\mathcal{G}) + E_{pa}(\mathcal{W},\mathcal{P}|\mathcal{G}),
    \end{aligned}
    \label{eqn:pm_full}
\end{equation}
where $E_{po}$ is energy term for pose stage updating the pose parameter $\theta$; $E_{sh}$ constrains the shape related parameters $\{\beta,\overline{\mathbf{T}},\mathcal{S}\}$, and $E_{pa}$ is for parameter stage, which update $\mathcal{W}$ and $\mathcal{P}$.
To avoid collision between muscles, bones and skins, 
we introduce a coupling penalty term $E_{cp}$ (Equation (\ref{eqn:ecol}) with $\lambda_{rep}=1$) to $E_{po}$, $E_{sh}$, and $E_{pa}$ throughout the training procedure, while assigning different weights to balance the impact on different stage.
We minimize $E_1$ by interactively going through the three stages until convergence. 

\paragraph{Pose stage.}
Given each $g_{ij}$, $\tilde{\mathbf{T}}$, $\tilde{\mathcal{J}}$, and $\tilde{\mathcal{W}}$, we solve for the specific pose parameter $\theta_{ij}$:
\begin{equation}
    E_{po}(\theta|\mathcal{G}) =\sum_{i\in S} \sum_{j\in P} w_{jt} E_{jt}^{ij} + E_{edge}^{ij} + E_{preg}^{ij} +w_{cp} E_{cp}^{ij}.
\end{equation}
Joint term $E_{jt}$ forces the posed template to match $g_{ij}$ by measuring the L2 distance between posed template joint and the target joint annotation. 
$E_{edge}$ measures the edge length difference between posed template and $g_{ij}$. Such term provides a good estimation of pose without knowing the subject specific shape. 
Since metacarpals have a limited range of motion according to \cite{wang2019hand,handanatomy}, we add a regularization term to prevent the metacarpal joints from having unrealistic rotations:
\begin{equation}
    E_{preg}^{ij} = \sum_{ij}||\mathcal{B}({\theta}_{ij})||_2,
\end{equation}
where $\mathcal{B}(\cdot)$ is a binary mask selecting only metacarpal joints.

\paragraph{Shape stage.}
We then update the shape related parameters ($\beta,\mathcal{S}$), the joint regressor $\mathcal{J}$, and the general template $\overline{\mathbf{T}}$ in this stage.
Firstly, we optimize the subject specific template $\mathbf{T}_p$ and ${J}_p$, which is directly relevant to $\overline{\mathbf{T}}$ and $\mathcal{J}$:
\begin{equation}
    E_{sh}(\mathbf{T}_p,{J}_p|\mathcal{G}) =  \sum_{i \in S} \sum_{j \in P} E_{geo}^{ij} + E_{reg}^{ij} + w_{jt}E_{jt}^{ij} + w_{jreg}E_{jreg}^{i} +w_{cp} E_{cp}^{ij},
\end{equation}
where $E_{geo}$ is a geometry term (Equation (\ref{eqn:geo})); $E_{reg}$ the regularization terms (Equation (\ref{eqn:reg})); $E_{jt}$ is a joint term; $E_{jreg}$ is a joint regularization term.
For the geometry term, We use a lower weight at the interior boundary of the muscle groups, namely the contacting vertices between each muscle, to ensure a consistent muscle boundary. Additionally, we use $E_{reg}$ on muscle vertices to ensure smoothness.
$E_{jreg}$ is a joint regularization to confine the joint locations of the subject $i$ consistent with joints prediction from the initial joint regressor:
\begin{equation}
    E_{jreg}^i =||\tilde{\mathcal{J}} \mathbf{T}_p^i- {J}_p^i||_2^2.
\end{equation}
After learned the $\mathbf{T}_p$ and ${J}_p$ by optimizing $E_{sh}(\mathbf{T}_p,{J}_p|\mathcal{G})$, we can get $\mathcal{J}$ by enforcing ${J}_p$ and $\mathcal{J}\mathbf{T}_p$ to be equivalent.
We then run principal component analysis (PCA) on $\mathbf{T}_p$ to obtain shape space parameters $\{\overline{\mathbf{T}}, \mathcal{S}, \beta\}$, where $\overline{\mathbf{T}}$ is the mean shape of MRI dataset $\mathbf{T}_p$. $\mathcal{S}$ is the principal component matrix, and $\beta$ is the PCA coefficient vector of the shape space.

\paragraph{Parameter stage.}
We optimize skinning weight $\mathcal{W}$ and pose blend shape $\mathcal{P}$ by: 
\begin{equation}
    E_{pa}(\mathcal{W}, \mathcal{P}|\mathcal{G}) = E_{geo} + E_{reg} + w_{jreg}E_{jreg} + w_{cp}E_{cp} + E_{wreg} + E_{pbreg}
\end{equation}
Similar to \cite{loper2015smpl}, $E_{pbreg}$ regularize the Frobenius norm of $\mathcal{P}$ to be zero, which prevents overfitting of the pose-dependent blend shapes. $E_{wreg}$ regulate the skinning weight by minimizing the distance between $\mathcal{W}$ and $\tilde{\mathcal{W}}$.

\paragraph{Pose Augmentation.}
After $E_1$ optimization, NIMBLE can potentially be directly used to better estimate hand pose because of more reliable bone and muscle modeling. As shown in Figure \ref{fig:mano_vis_compare}, we are able to provide anatomically correct and physically plausible deformation compared with state-of-the-arts.
However, due to limited hand poses provided by the MRI datasets \cite{li2021piano,wang2019hand}, NIMBLE may suffer from degraded performance in applications requiring large hand pose variances. To address this issue, we additionally optimize NIMBLE on hand scans from MANO dataset \cite{romero2017embodied}.
MANO provides 1554 raw scans and hand registrations align with the topology of MANO hand model. We perform a topology transfer with a simplified physically based non-rigid registration (Section \ref{sec:registration}) to inline our model to MANO dataset.
To achieve this, we compute a dense correspondence from MANO topology to ours by manually fitting MANO to our template using Wrap3D \cite{webr3ds}.
Then we run the non-rigid optimization with geometry term, non-rigid elasticity term and dense correspondence to match with raw scan. 
By doing so, we obtain another 1554 hand registrations with large pose variance.

We follow the same parameter learning strategy as in the MRI dataset and further optimize $\mathcal{G}$ learnt from MRI scans. 
Note that the canonicalized MANO data only contains skin geometry, resulting in weak supervision on bone and muscle. 
To prevent unexpected deformations, we leverage an additional shape regularizer to constrain the deformation of inner geometry. Essentially, we want to use the skin to guide the deformation of bone and muscle so that the inner and outer mesh will not downgrade to the average template in our previously registered MRI shape space.
We define the shape regularizer as follows:
\begin{equation}
    E_{sreg} = \sum_i||\mathbf{T}_p^i\mathcal{S}^T||^2, 
\end{equation}
where $\mathbf{T}_p^i\mathcal{S}^T$ denotes the projected shape coefficients ${\beta}_i$ on to the MRI shape space, which corresponds to ${\beta}_i = {0}$.

More parameter setting, registration and learning details can be found in Section.~\ref{sec:details}.

\section{Photorealistic Rendering}
\label{sec:rendering}
Modeling the high-quality and realistic appearance is important for a realistic rendering pipeline. 
Physically-based textures, including diffuse albedo, normal maps, specular maps, play an important role in rendering photo-realistic hand appearance. 
Here, we introduce how to model the appearance for NIMBLE, \textit{i.e.,} $\mathcal{A}(\alpha)$. 
\\

\paragraph{Appearance Capture.}
We utilize a photometric appearance capture system that we call HandStage analogous to the USC LightStage \cite{debevec2000acquiring} to capture the detailed physically-based textures.
We are able to attain diffuse albedo, normal maps and specular maps of hands by applying several patterns of polarized gradient illumination on them.
We captured 20 hands of different identities with our HandStage capture system to reconstruct 8192x8192 pore-level detailed physically-based hand textures.
To increase diversity, we include extra 18 online hand texture assets from \cite{webscan3d}.
Our final appearance dataset consists of 38 photo-realistic hand texture assets from different ages, genders and races.

For rich and authentic hand appearance generation, we create a parametric appearance model from our appearance dataset.
Every asset in our dataset have physically-based textures as well as uniform texture UV mapping, which allows us to apply linear interpolation between existing textures.
For every appearance $\mathbf{A}_i$ in our dataset, we compute $\mathbf{\bar A}=\frac{1}{n}\sum_{i=1}^n \mathbf{A}_i,n=38$ as the average appearance, including average diffuse albedo, normal maps and specular maps.
Then, we run principal component analysis (PCA) using singular value decomposition to obtain the principal components $\Phi$ from every existing appearance in our total appearance dataset.
After that we obtain the parametric appearance model for appearance parameter vector $\alpha \in \mathbb{R}^k$ as
\begin{equation}
    \mathcal{A(\alpha)} = \mathbf{\bar A}+\Phi\alpha,
\end{equation}
where $k$ is the number of principal components, here we choose $k=10$.
With parametric appearance model created from PCA, we could also generate realistic physically-based textures out of our dataset.
Since our textures have uniform texture UV mapping as our template skin mesh, we could directly apply generated physically-based textures NIMBLE with different shapes and produce a photo-realistic appearance.
\\

\begin{figure}
    \centering
    \includegraphics[width=0.9\linewidth]{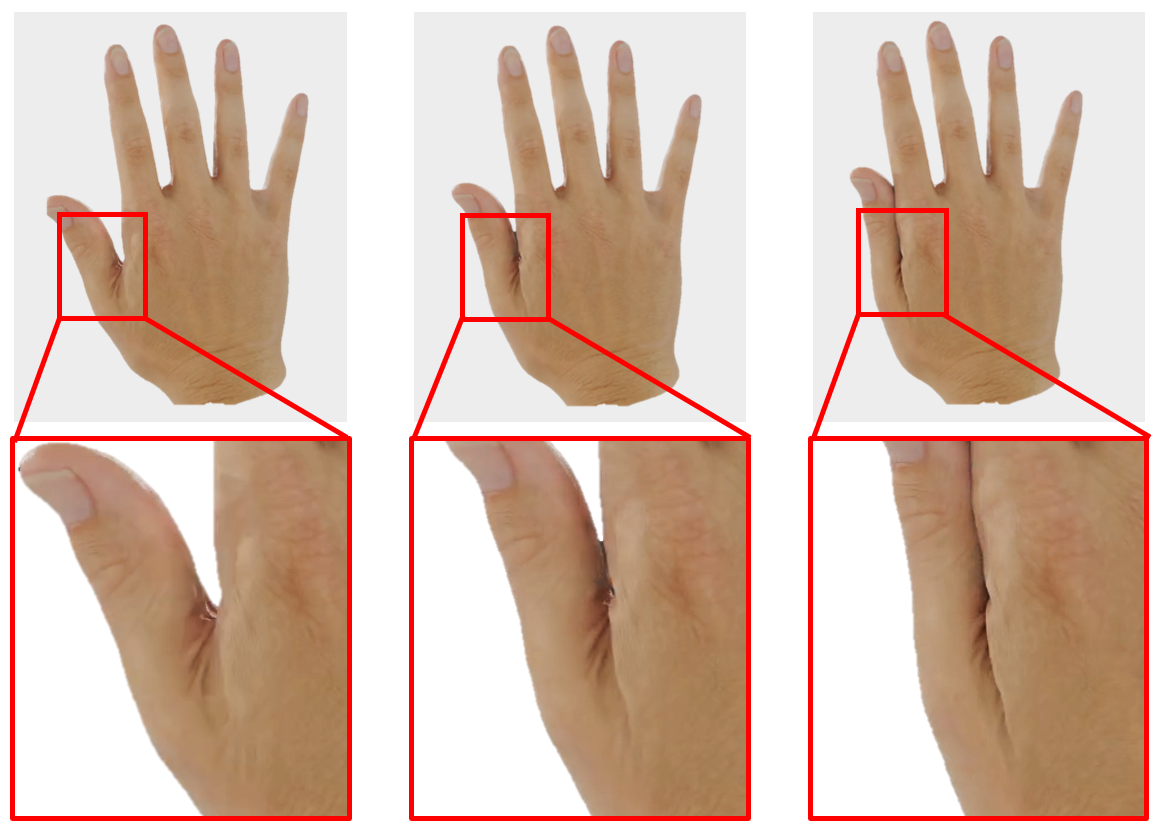}
    \caption{Simulation of skin wrinkle due to thumb adduction. Note the wrinkle gradually appear and skin bulges as the fingers become closer.}
    \label{fig:wrinkle}
\end{figure}

\begin{figure*}
    \centering
    \includegraphics[width=\linewidth]{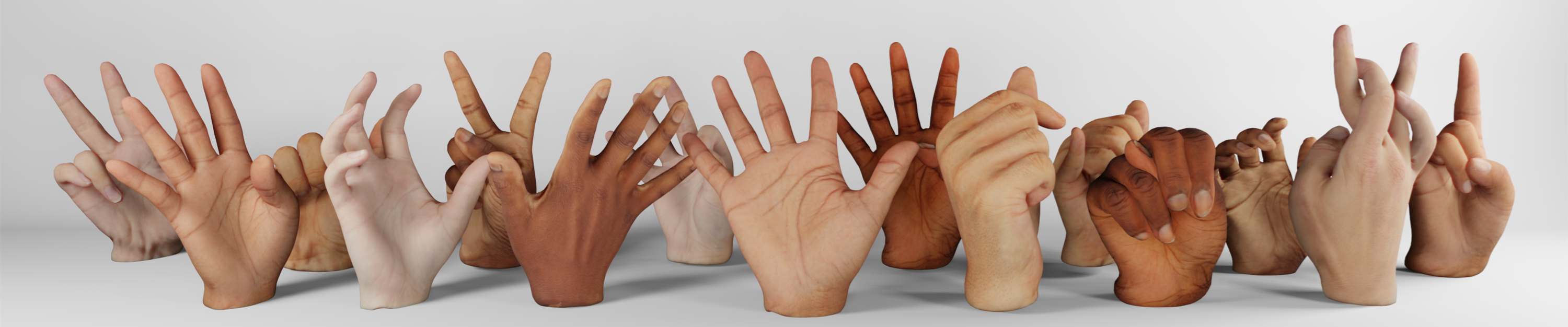}
    \caption{Gallery of generated 3D Hand. Our model is able to synthesize realistic digital hand with large variance on pose, shape and appearance. }
    \label{fig:gallery}
\end{figure*}

\paragraph{Rendering Process.}
\begin{figure*}
    \centering
    \includegraphics[width=\linewidth]{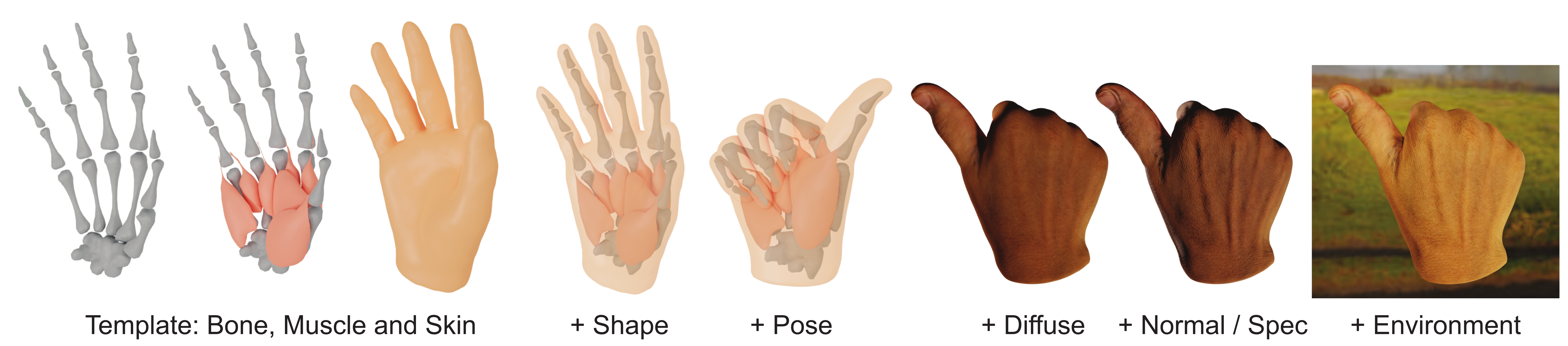}
    \caption{Our rendering pipeline for creating photorealistic hand. Starting with our full template, we randomly generate shape, pose and appearance and render the 3D hand with environment maps. We also show difference rending effects with only diffuse map and with normal map.}
    \label{fig:renderprocess}
\end{figure*}
We show our procedural rendering process in Figure \ref{fig:renderprocess}, we start with NIMBLE template model, and generate shape, pose and appearance parameters randomly. 
NIMBLE takes the parameters as input and generates a realistic 3D hand with bone, muscle and skin geometry, as well as photo-realistic appearance with diffuse map, normal map and specular map. 
Then we render the 3D hand with Cycles, a photo-realistic ray-tracing renderer \cite{cycles}.
We employ image-based lighting with high dynamic range images (HDRI) as background texture to illuminate the hand, as shown in Figure \ref{fig:renderprocess}.
We also show a variety of different hand poses, shapes and appearances under a uniform lighting in Figure \ref{fig:gallery}. 
To generate a dynamic motion sequence, we map pose parameters to full pose quaternion representation, and linearly interpolate between different pose to keep a smooth pose morphing. Please see the supplementary video for examples.
To further enhance our rendering quality, we generate skin wrinkles with cloth simulation in rendering engines. 
Here we only use the surface triangles of our volumetric mesh so that cloth simulation schemes can be applied.
See Figure \ref{fig:wrinkle} for an example.
During the adduction of the thumb to the index finger, the first dorsal interossei between thumb and index finger contract and pull the bones closer, the contracted muscle makes the skin bulge, and the purlicue skin is then squeezed to create a fold. 
As can be seen in the close-up of Figure \ref{fig:wrinkle}, the wrinkle near purlicue gradually appears as the fingers become closer. NIMBLE recovers muscle and skin bulging under such pose, while cloth simulation produces wrinkles caused by the squeezing between the geometry of thumb and index finger.

\section{Experimental Results}
\label{sec:experiment}
We first show our implementation details, then evaluate the performance of our registration process and the NIMBLE models learned from these registrations. 
We test NIMBLE on public hand datasets and compare with state-of-the-arts, \textit{e.g.,} MANO \cite{romero2017embodied} and HTML \cite{qian2020html}. Extensive experiments demonstrate NIMBLE's outperforming performance and capability of photo-realistic rendering among diverse datasets. 
Additionally, we show that NIMBLE can be easily fitted into the hand inference pipeline from various input.

\subsection{Implementation Details}
\label{sec:details}
%
In our experiments, we set the term weights in Equation (\ref{eqn:reg_full}) according to template type during registration. For muscle, since the target is not separated, we use larger regularization term weights to guide the registration. We set $\lambda_d = 1$, $\lambda_{dn} = 1$, $\lambda_a = 0.1$, $\lambda_e = 0.5$, $\lambda_{fn} = 0.01$, $\lambda_{ne} = 0.02$, $\lambda_{rep} = 0.5$, $\lambda_{col} = 0.1$ and $\lambda_{att} = 10$.
For skin, we set larger geometry term weights and non-rigid elasticity term weights: $\lambda_d = 2$, $\lambda_{dn} = 2$, $\lambda_{ne} = 1$, $\lambda_{col} = 1$,  $\lambda_{a} = 0.01$.
Specifically, for bone, we omit the non-rigid elasticity term with $\lambda_{ne} = 0$.
The whole registration takes approximately 20 minutes for the muscle group, 8 minutes for skin, and 2 minutes for bone. 
For parameter learning, we also use weights to balance each term. During the first iteration, we set the weight $w_jt$ and $w_{jreg}$ in Equation (\ref{eqn:pm_full}) to 100 to enforce a strong pose constraint, and decrease to them to 0.01 and 10 respectively after the first stage. We set the weight $w_{cp}$ for coupling penalty gradually increase from 0.1 to 1 to ensure that our final model is collision-free. 
We also adopt an additional optimization process with Equation (\ref{eqn:icol}) and (\ref{eqn:ecol}) to handle skin collisions with bone and muscles during model usage.
We iterate the whole process of registration and parameter learning several times to get a stable result. All of our experiments are performed with PyTorch auto differentiation on an NVIDIA GeForce RTX 3090.

\subsection{Registration Evaluation}    

\begin{figure*}
    \centering
    \includegraphics[width=\linewidth]{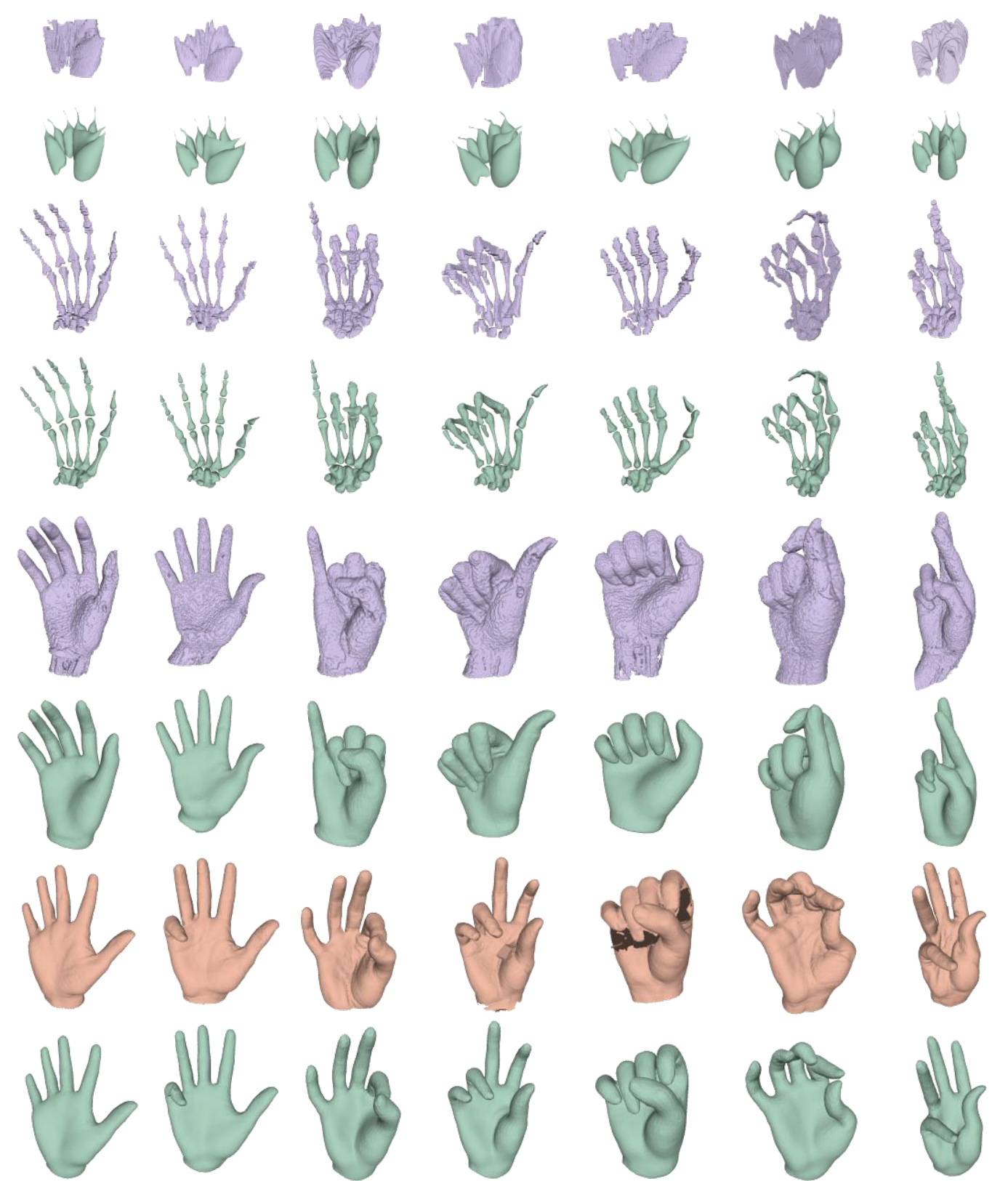}
    \caption{Qualitative result of MRI registration. From top to bottom: MRI muscle mesh (purple) and our registration (green), MRI bone mesh and registration, MRI skin mesh and our registration, as well as the MANO scan mesh (yellow) and our registration.}
    \label{fig:reg_quali}
\end{figure*}

\paragraph{Quantitative Result.}
\begin{figure}
    \centering
    \includegraphics[width=\linewidth]{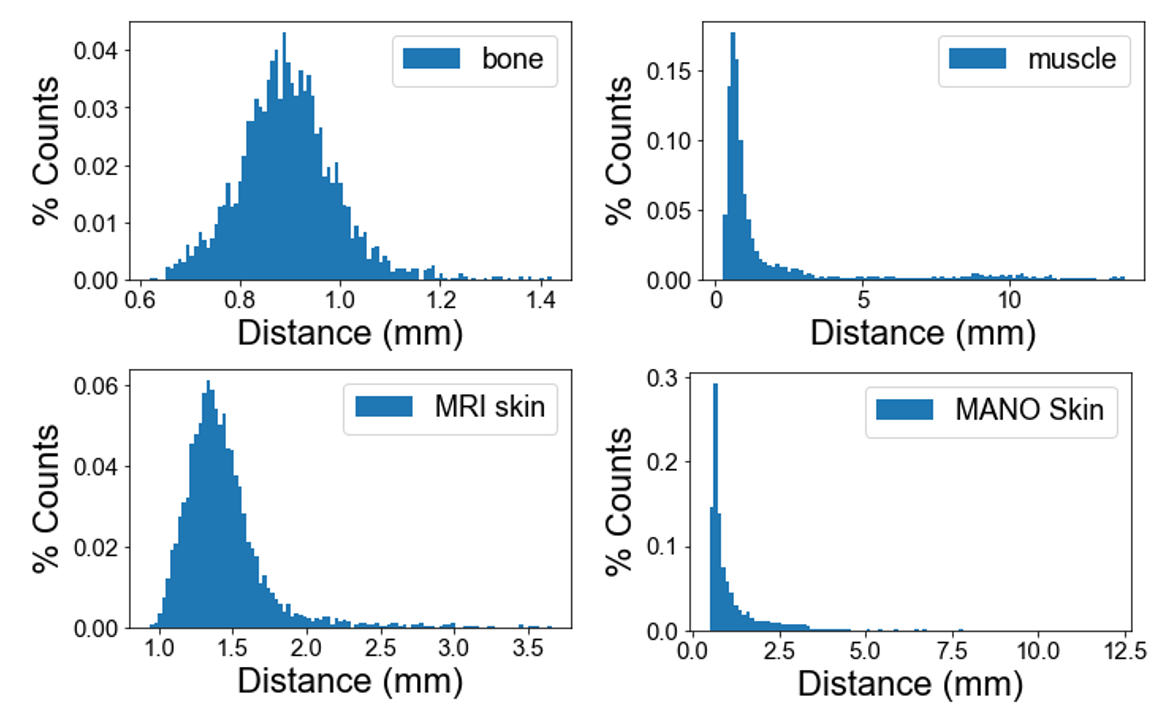}
    \caption{The histogram of median per-vertex distance between registration and the target surface.}
    \label{fig:reg_eval}
\end{figure}
In Figure \ref{fig:reg_eval}, we plot histograms for the median of per-vertex distance from registration to the MRI mesh and MANO scan mesh. The distance is measured across all registered MRI data and MANO scans. 
Note that we discard the inner vertices in this evaluation.
It is indicated that our method produces registrations that generally match the target mesh within a 2 mm distance error. 
For bone registration, almost all vertices have a distance error below 1.4 mm, while 87\% of the vertices achieve a median distance less than 1 mm.
For muscle registration, 65\% of the vertices make a median distance less than 1 mm and only 9\% vertices are above 5 mm. These are caused mainly by the missing data at the attachment part where muscle gets thinner and attaches to bone, as described in Section \ref{sec:data_capture}. We thus perform higher regularization weights and attachment terms to increase robustness at these parts.
For skin registration in the MRI dataset, 70\% of the vertices make a median distance less than 1.5 mm, while the max distance error is 3.69 mm. 
While for the MANO scan skin dataset, with our topology transfer method, we achieve a mean error of 1.09 mm. There are 4\% of vertices error above 3 mm. This is mostly caused by incomplete scan in the dataset, especially with object occlusion, as shown in Figure \ref{fig:reg_quali} (last two rows). 
Meanwhile, we report the mean squared distance from scan to registration for MANO skin is 1.76 mm. 

\paragraph{Qualitative Result.}
Figure \ref{fig:reg_quali} shows representative registrations sample of the MRI dataset and MANO scan dataset. 
For all inner and outer data, our method provides an accurate and smooth registration. We are able to capture the muscle stretching and bulging effects. 
It is notable that our registration is able to maintain a robust performance towards noisy MRI target while capturing detail skin wrinkles from detailed MANO scan \cite{romero2017embodied}. \\

\paragraph{Ablation Study.}
\begin{figure}
    \centering
    \includegraphics[width=\linewidth]{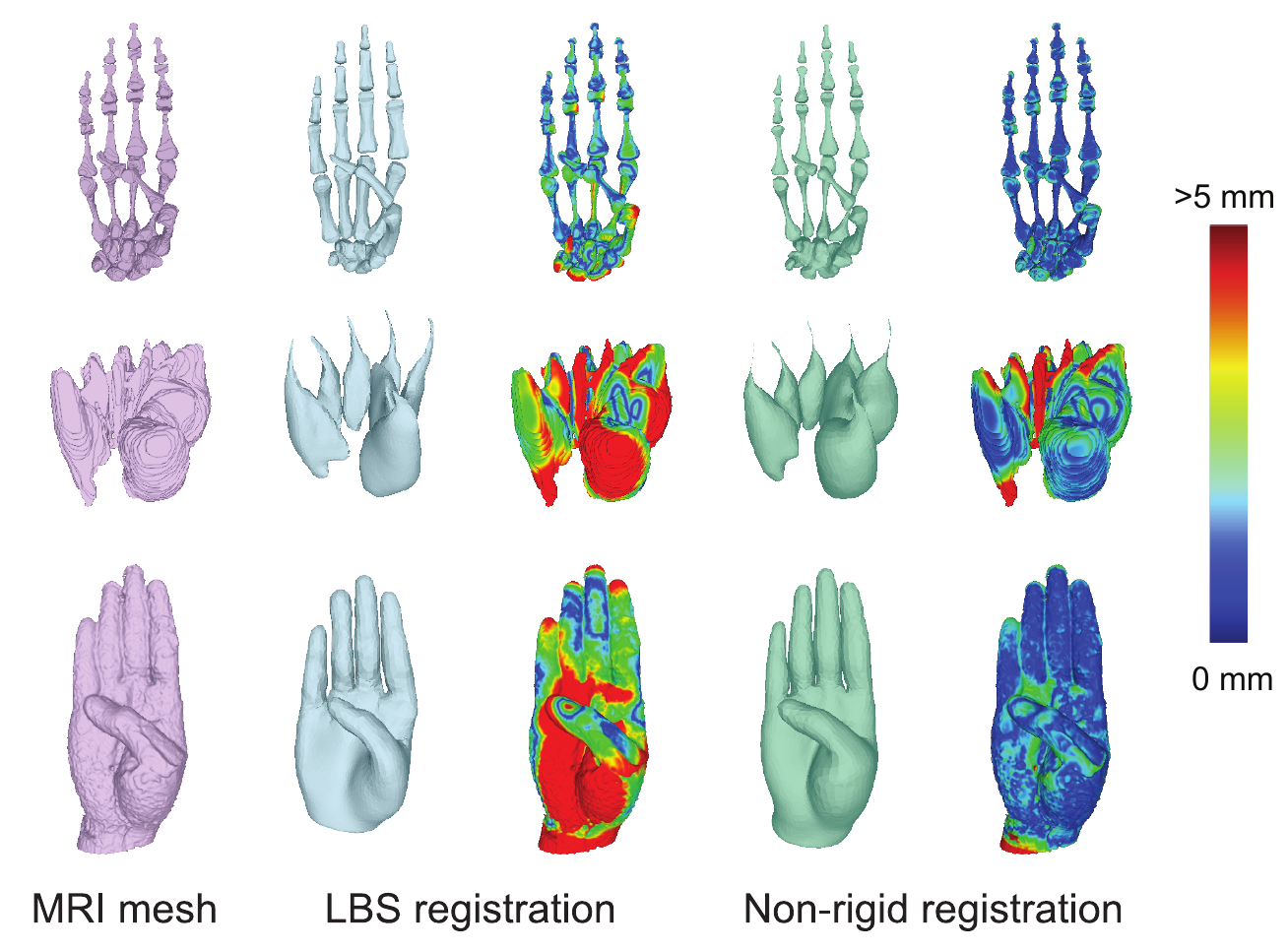}
    \caption{Results of the LBS initialization and non-rigid refinement on MRI mesh. From left to right: target MRI mesh, LBS registration and mri to registration distance, Non-rigid registration and the color-coded distance. }
    \label{fig:reg_process}
\end{figure}
Our registration process contains two steps: LBS pose initialization and iterative non-rigid refinement. 
Figure \ref{fig:reg_process} visualized the registration results of each optimization step. 
The LBS step serves only as a pose initialization, it is unable to capture the details of muscle and skin bulging, especially around the thumb muscle and hand palm.
After the iterative non-rigid refinement, the registration tightly fits the surface of the MRI mesh. 
Note on the top right of Figure \ref{fig:reg_process}, the geometry error remains large in the middle of the MRI muscle mesh. This is due to segmentation error, some tendons are mislabeled as muscle because it is hard to distinguish them on MRI slices. 
While using the strong regularization term in Section \ref{sec:registration}, we successfully match the template to the target without fitting it to mislabeled tendons.

\begin{figure}
    \centering
    \includegraphics[width=\linewidth]{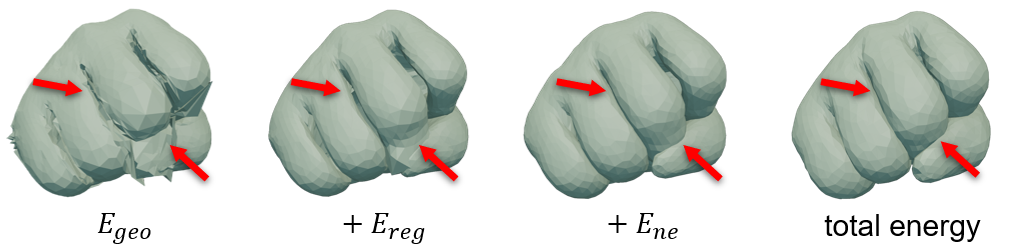}
    \caption{Registration results with different optimization terms. We label regions of interest with arrows. From left to right, we run registration with geometry term and then add regularization term, non-rigid elasticity term, and the collision terms to obtain the total energy.}
    \label{fig:reg_ablation}
\end{figure}

To assess the impact of each of our energy terms on iterative refinement, we register an MRI hand mesh with multiple energy term variants.
As illustrated in Figure \ref{fig:reg_ablation}, we begin with the geometry term and then add the regularization, non-rigid elasticity, and collision terms individually to obtain the total energy defined in Section \ref{sec:registration}.
While the geometry term forces the vertex to align with the target, the fingers collide with one another, resulting in severe artifacts.
While adding a regularization term eliminates some artifacts, finger collision and unnatural knuckle deformation remain.
With the addition of non-rigid elasticity term, the deformations of each finger become more realistic, but the thumb and middle finger continue to self-penetrate.
By including a collision term, the self-penetration problem is resolved and the final collision-free registration result is obtained.

\subsection{Model Evaluation}

The general method to evaluate a statistical model is to measure its compactness and generalization~\cite{romero2017embodied,loper2015smpl}. Compactness measures the amount of variability in the training data captured by the model, while generalization measures the model ability to represent unseen shape, pose and appearance.

\begin{figure*}
    \centering
    \includegraphics[width=\linewidth]{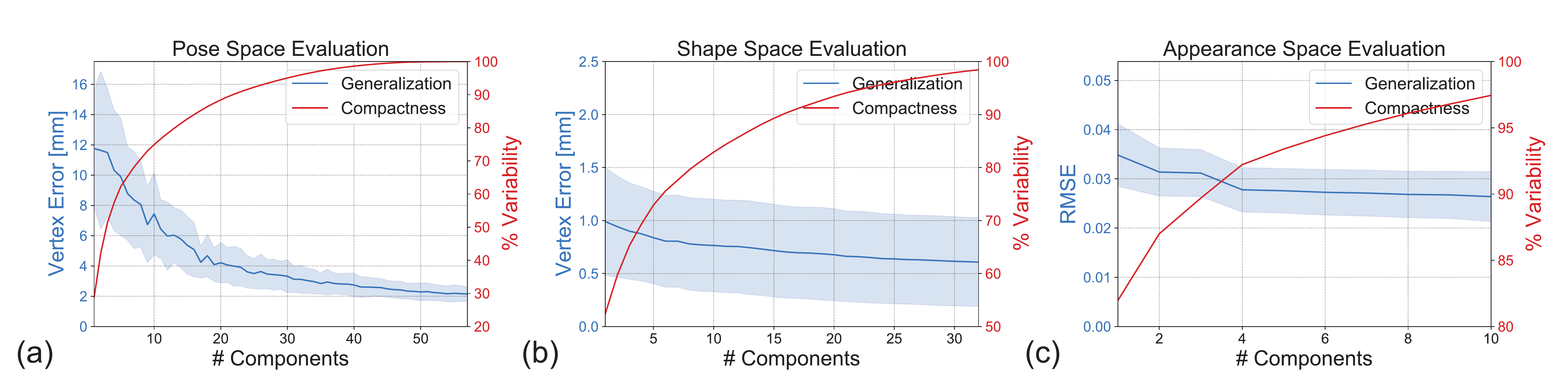}
    \caption{Model Quality of compactness and generalization. (a) Pose space (b) Shape space (c) Appearance space.}
    \label{fig:pm_eval}
\end{figure*}

\begin{figure}
        \centering
        \includegraphics[width=\linewidth]{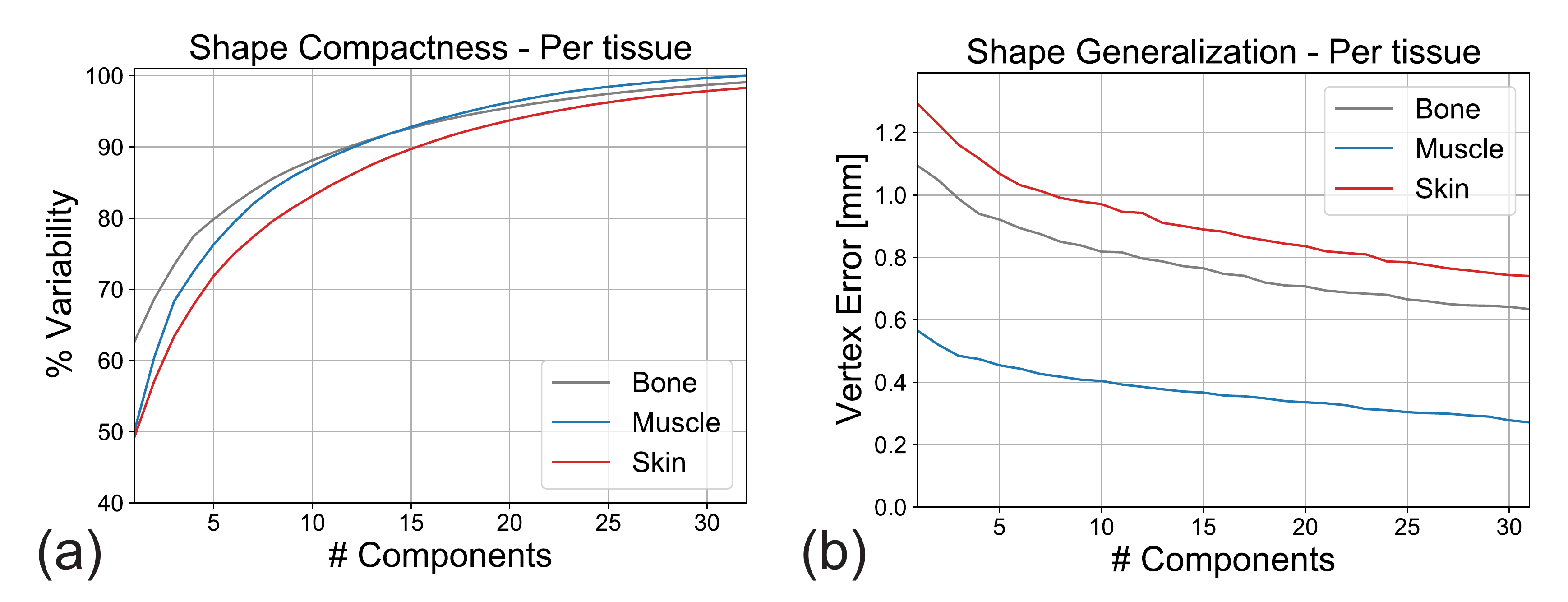}
        \caption{Per-tissue shape compactness and generalization.}
        \label{fig:pm_eval_pt}
    \end{figure}

\paragraph{Compactness.}
Figure \ref{fig:pm_eval} (a), (b) (red curve) plot the compactness of the NIMBLE shape and pose space, respectively.
These curves depict the variance in the training data captured by a varying number of principal components. 
The pose space plot shows that 15, 25, 30 components can express 83\%, 92\%, 95\% of the full space. 
The result is consistent with the anatomy of the human hand, which is generally considered to have 27 degrees of freedom \cite{handanatomy}.
As for the shape space, 
as shown in Figure \ref{fig:pm_eval} (b), we can note that the first principal component covers 50\% of the shape space. Meanwhile, 10 and 20 components manage to cover 83\% and 93\% of the complete shape space.
We also plot the per-tissue compactness curve in Figure \ref{fig:pm_eval_pt}(a). It indicates that the variance of different tissues is mostly consistent. 

\paragraph{Generalization.}
To study the generalization ability of NIMBLE shape space in the presence of limited shape variance, we perform a leave-one-out evaluation on our MRI and MANO training set, which contain 62 individuals in total. 
Figure \ref{fig:pm_eval}(b) blue curve shows the generalization curve of the shape space. 
We report the mean squared error and the standard deviation in millimeters.
As the number of principal components increases, the mean error decreases to a minimum error of 0.6 mm achieved by the full space.
We also plots the per-tissue generalization in Figure \ref{fig:pm_eval_pt}(b). 
Note that the muscle error is the lowest across all components, meaning the shape variance for muscle is relatively small compared to bone and skin.

To evaluate the generalization capabilities of the pose space in NIMBLE, namely the ability to generalize to unseen pose with known shape parameters.
We construct a test set containing 9 registered MRI data in the unseen pose, combined with the test scan set from \cite{romero2017embodied}.
The test scan set contains 50 hand surface scans with unseen poses and shapes.
All meshes are in alignment of our topology and none were used to train our models. 
We fit our trained model to the registered mesh, optimizing over pose $\vec{\theta}$ and $\vec{\beta}$ to find the best fit in terms of the mean squared vertex distances. Since we are evaluating pose generalization ability, we use full shape space for this experiment.
Figure \ref{fig:pm_eval}(a) blue curve shows the generalization results.  
We report the mean squared distance error and the standard deviation.
Similar to shape space, this plot for the pose space decreases monotonically for an increasing number of components.

\begin{figure}
        \centering
        \includegraphics[width=0.85\linewidth]{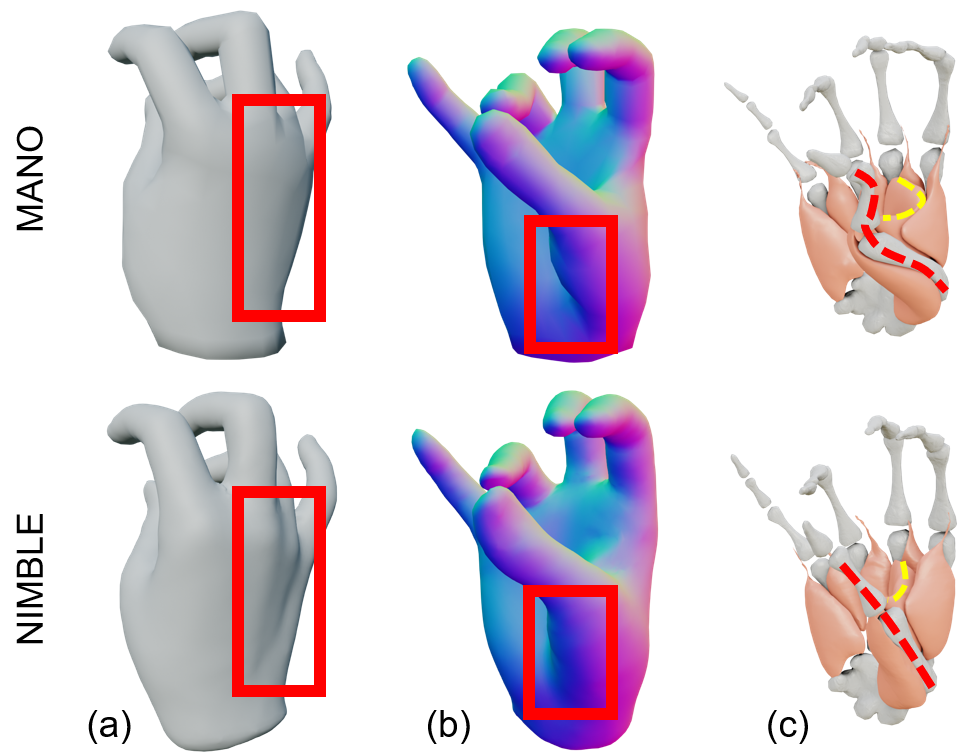}
        \caption{Deformation comparison with MANO \cite{romero2017embodied}. (a) NIMBLE retains skin details during deformation, while MANO provides an overly smoothed skin. (b) (c) MANO presents implausible flexion of the inner bone and muscle, as well as an unrealistically sunken skin, whereas NIMBLE maintains anatomically correct and physically plausible deformation.}
        \label{fig:mano_vis_compare}
    \end{figure}

\paragraph{Compare with MANO.}
To compare with MANO \cite{romero2017embodied}, similar to the pose generalization experiment, we fit models to our MRI test set and their scan test set, respectively.
We use full pose and shape space for all models in this experiment.
In Table \ref{tab:pm_pose_gen}, we report the mean squared vertex error in millimeters.
It can be seen that MANO model performs best in its test set but does not generalize well to MRI data.
Meanwhile, our model is not able to generalize to unseen pose and shape in the MANO test set with low pose variance derived from MRI data.
After pose augmentation, though the performance on MRI test set drops a little bit, the result on MANO test set is significantly improved. 
Overall, our model achieves satisfying on both test sets and achieves the smallest average error. 
Figure \ref{fig:mano_vis_compare} further shows qualitative hand fitting comparison with MANO. It is notable that MANO suffers from impractical inner deformation and lacks skin details, as it is built on outer surface only. 
In contrast, our NIMBLE model achieves anatomically correct inner hand tissue deformation while retaining skin details.
\begin{table}[]
    \caption{Comparison with MANO. We evaluate mean squared distance in millimeters on MRI test set and MANO test set respectively, and report the average error.}
    \begin{tabular}{l|c|c||c}
    \hline
    Model             & MRI test & MANO test & Avg.          \\ \hline \hline
    MANO \cite{romero2017embodied}              & 3.32             & \textbf{1.46}           & 2.39             \\ \hline
    Ours - MRI        & \textbf{2.51}    & 3.89                    & 3.20                     \\ \hline
    Ours - Pose Aug  & 2.67             & 1.62                    & \textbf{2.15}            \\ \hline
    \end{tabular}
    \label{tab:pm_pose_gen}
    \end{table}
    
\paragraph{Bone/Muscle/Skin Correlation.}
Regarding the correlation of bone, muscle and skin, we impose biomechanical constraints to correlate three tissues implicitly encoded via the non-rigid elasticity term in registration (Section \ref{sec:registration}) and the coupling term in parameter training (Section \ref{sec:pm_train}). These terms force skin deformations to follow bone and muscle movements, critical for physically correct simulations. 
Visually, such deformations are more nuanced, as shown in Figure \ref{fig:mano_vis_compare} and the accompanying video. They both demonstrate the bulging thumb base (thenar eminence) when the thumb touches the index finger, illustrating the intricate coordination between bones, muscles, and skin.
To assess the impact of inner muscle layer quantitatively, we conduct an ablative study on bone-skin vs. bone-muscle-skin models. We fit the MRI test set and MANO test set with models learned on bone-skin and bone-muscle-skin data separately using full pose and shape space. The evaluation results are shown in Table \ref{tab:model_ablation}. 
It can be seen that adding muscle layer achieves lower error on all metrics. Compared to bone, skin error shows larger improvement on both test sets, indicating that muscle layer has a positive impact on skin deformation. We thus conclude that modeling muscle layer facilitates both visual realism and fitting results, and the correlation between each tissue is successfully encoded in the model through our registration and parameter learning pipeline.

\begin{table}[]
    \caption{Ablative comparison of bone-skin model and bone-muscle-skin model. We evaluate mean squared distance in millimeters on bone and skin mesh from MRI test set, as well as skin mesh from MANO test set.}
    \begin{tabular}{l|c|c|c}
    \hline
    Model             & MRI-bone & MRI-skin & MANO-skin  \\ \hline \hline
    Bone-Skin         & 2.59     & 2.61     & 1.67       \\ \hline
    Bone-Muscle-Skin  & \textbf{2.58}     & \textbf{2.56}     & \textbf{1.62} \\ \hline
    \end{tabular}
    \label{tab:model_ablation}
    \end{table}

\begin{figure}
    \centering
    \includegraphics[width=\linewidth]{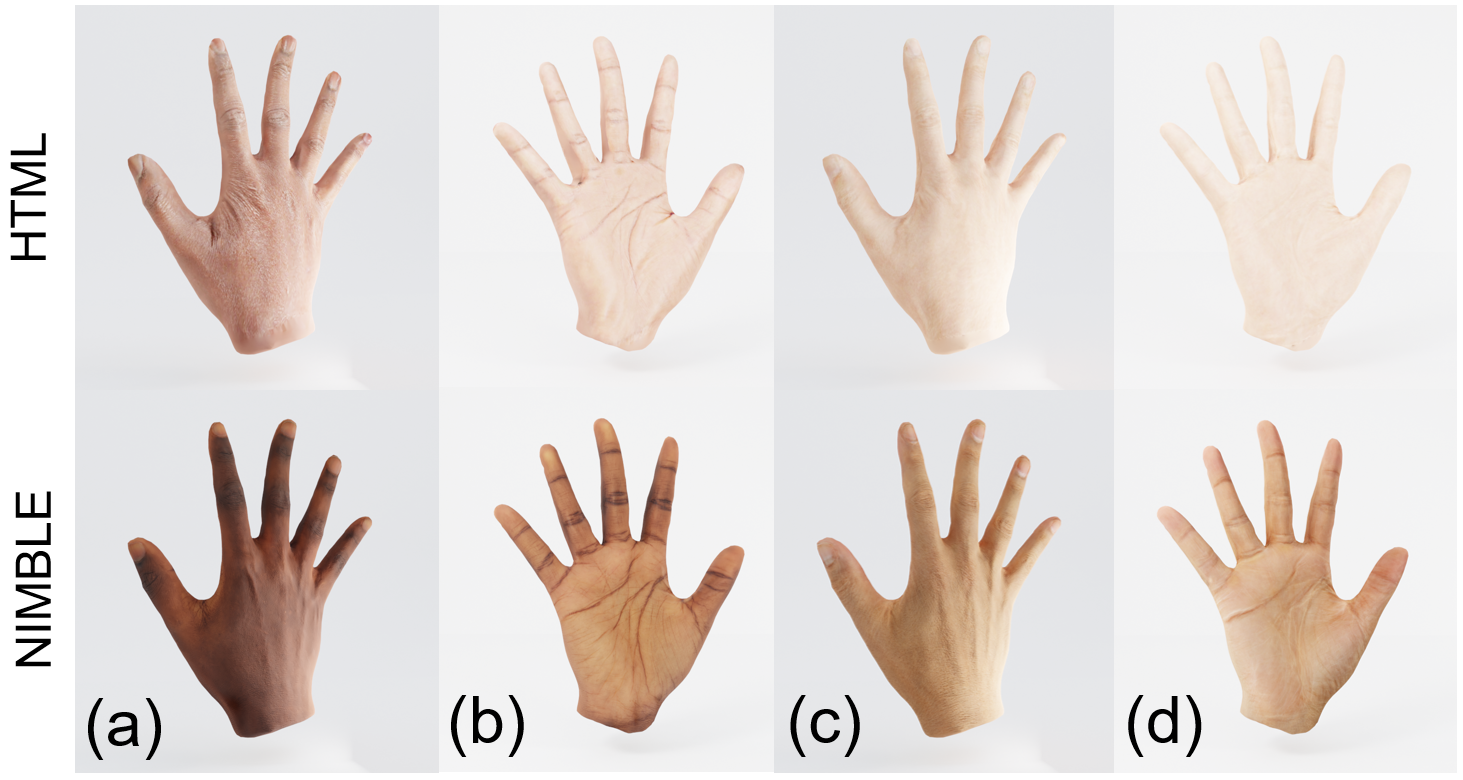}
    \caption{Appearance comparison with HTML \cite{qian2020html}. 
    We render our model with textures from HTML and our appearance data, respectively. (a)(b) shows the back and front side of captured hand textures from each model, (c)(d) shows random sampled textures. (a)(c) are  rendered with an additional lighting source to highlight the normal differences.}
    \label{fig:tex_compare}
\end{figure}

\paragraph{Photorealistic Appearance.}
Figure \ref{fig:pm_eval}(c) shows the evaluation of our appearance model.
The plot depicts the rising variability in our appearance dataset as the number of employed principal components increases.
The first several components could represent a significant amount of variation which mainly include skin tone and ruddiness, while the other components control the details of skin.
For evaluating generalization, we perform a leave-one-out evaluation similarly. 
Since our appearance dataset includes diffuse albedo, normal maps and specular maps, we utilize root mean squared error (RMSE) as the metrics for reconstruction error measurement.
We reconstruct the left-out textures using the PCA analyzed from the other textures and measured the reconstruction error as RMSE of the vectorized textures. 
As shown in \ref{fig:pm_eval}(c), the reconstruction error decreases as the number of components increases.
Figure \ref{fig:tex_compare} shows a qualitative appearance comparison with HTML \cite{qian2020html}. 
We use Wrap3d \cite{webr3ds} to transfer texture from HTML to our model and render the result under the same lighting condition. 
As can be seen that the appearance submodule of NIMBLE covers a wide diversity of skin complexions. In particular, the use of normal map in NIMBLE better illustrates tendons on the back of the hand and palm lines despite complexion variations.

\subsection{Applications}

\begin{figure}
    \centering
    \includegraphics[width=\linewidth]{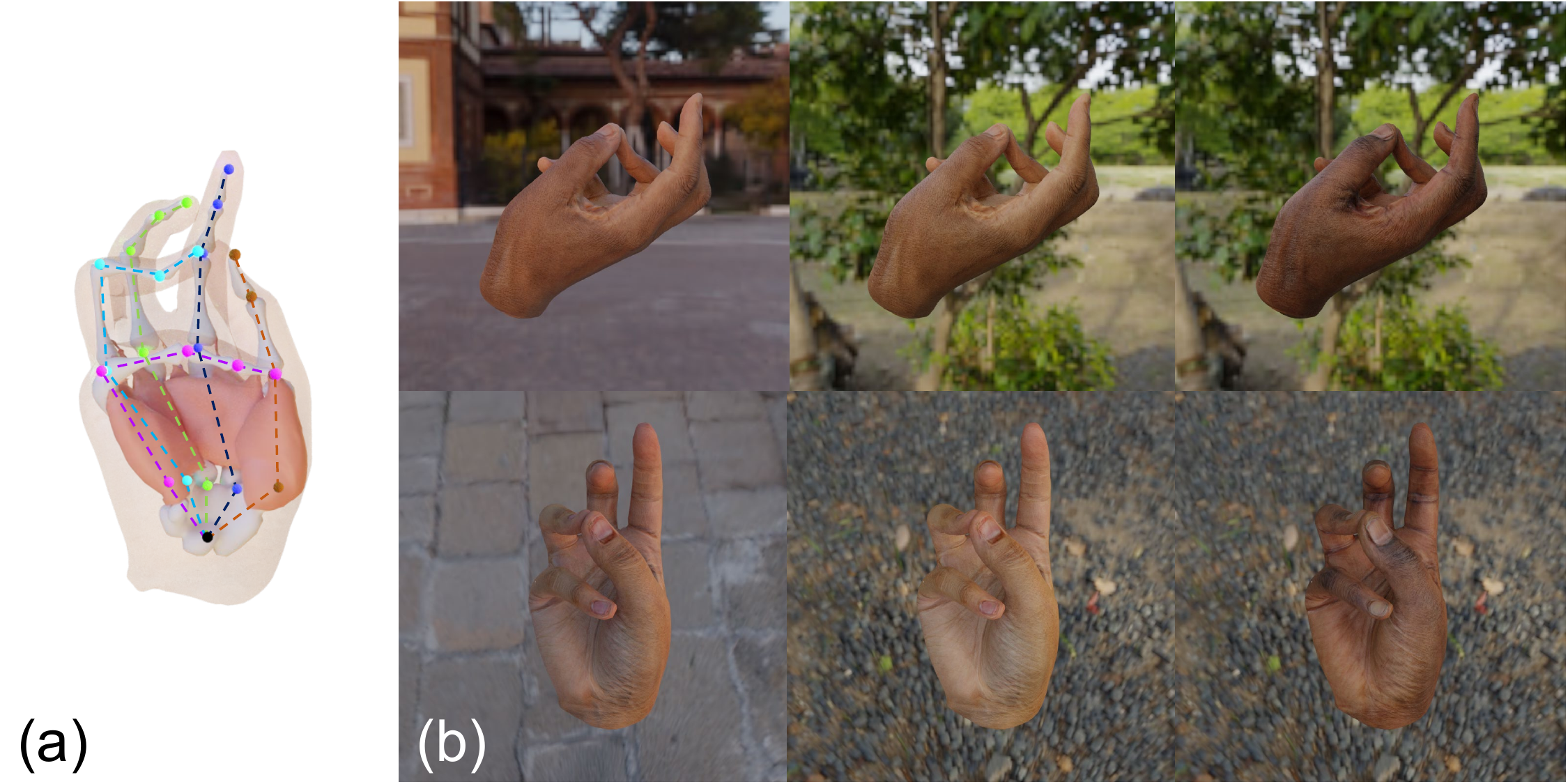}
    \caption{Representative results of the same posed hand with different camera view, illumination and texture. (a) Inner and outer geometry of NIMBLE generated digital hand and the corresponding 3D joint annotation. (b) Photorealistic images. From left to right, the first two columns show the same texture under different illumination, while the second and third column show different textures under the same illumination.}
    \label{fig:renderdataset}
\end{figure}

\paragraph{Synthesizing Digital Hand.}
Learning-based hand-related tasks rely on high quality labeled datasets of hand images, yet acquiring such datasets with correct labels (e.g. 3D geometry, pose and appearance) is extremely challenging owing to the high degree of freedom (DoF) of hand motions.
As each finger can flex and extend, abduct and adduct, and also circumduct; and all fingers can move independently as well as coordinately to form specific gestures. Such high DoF causes complex occlusions and hence imposes significant difficulties in skeleton annotation. Even for humans, it would be very difficult to manually label hand joints of complex gestures at high precision, largely due to the ambiguity caused by occlusions.
Our model is well-suited to help resolve these issues. With the NIMBLE model and render engines, we can create an unlimited number of photo-realistic hand images and video sequences with corresponding ground-truth inner and outer geometry, pose and texture maps. All of which can be used for downstream hand-related learning tasks. 
We demonstrate qualitative results of our photorealistic rendering and its ability to generate a complete digital hand in Figure \ref{fig:gallery}. 
Several results are shown in Figure \ref{fig:renderdataset} for the same pose with different texture under different lighting environments. We can also provide the corresponding ground truth 3D joint annotation. 
\\

\paragraph{Hand Inference.}
Like other parametric hand models \cite{li2021piano,romero2017embodied}, NIMBLE is easily adaptable to a variety of optimization and regression-based hand inference tasks, such as hand anatomy analysis, pose and shape estimation, and hand tracking with a variety of inputs including meshes, point clouds, MRI, and RGB images.
We integrate NIMBLE as a differentiable layer that takes shape $\theta$, pose $\beta$, and appearance $\alpha$ as input and outputs a 3D hand mesh with photorealistic textures and 3D joints. 
Similar to \cite{li2021piano,hasson2019learning}, NIMBLE support training with multiple losses, such as parameter loss, regularization loss, 2D/3D joint loss, mesh loss, as well as photometric loss \cite{qian2020html} and silhouette loss \cite{Xu:2018:MHP:3191713.3181973} with differentiable renderers provided in PyTorch3D \cite{ravi2020pytorch3d}.

\begin{figure}
    \centering
    \includegraphics[width=\linewidth]{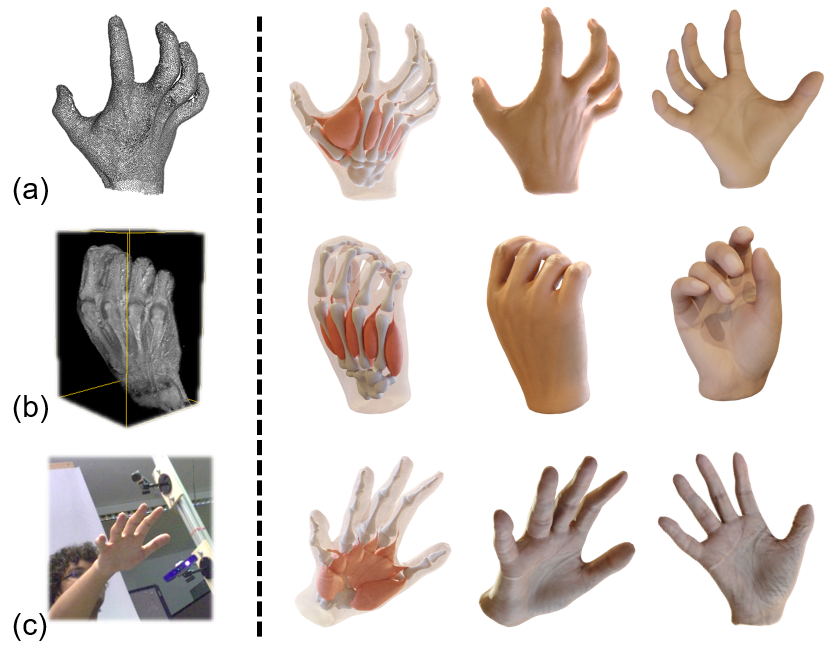}
    \caption{Representative results of the usage of NIMBLE for hand pose and shape estimation from (a) point cloud, (b) MRI volume and (c) RGB image. For textureless point cloud and MRI, NIMBLE is rendered with mean texture.}
    \label{fig:application_result}
\end{figure}

We show representative results for the usage of NIMBLE in Figure \ref{fig:application_result}. We are able to estimate and recover anatomically correct inner and outer hand structure and provide a photorealistic rendering from various inputs.
Figure \ref{fig:application_result} (a) shows an example of hand inference from point cloud. The input point cloud is taken from the MANO test set, and we perform this task using an optimization-based method \cite{newcombe2015dynamicfusion}. 
We optimize pose and shape parameters with joint loss and mesh loss with respect to the target point cloud. 
Figure \ref{fig:application_result} (b) illustrates the inference of hand anatomy from an MRI volume. We build a network with ResNet3D \cite{tran2018closer} encoder and a parameter regressor branch to directly regress NIMBLE parameters from MRI volume. We train the network on our MRI training set with supervision on the pose and shape parameters as we acquire ground truth parameter labels from our registration and parameter learning stages.
Since there is no appearance guidance in point cloud and MRI volume, we omit appearance parameter $\alpha$ during optimization and training and use the mean texture for rendering in Figure \ref{fig:application_result} (a)(b). 
Similarly, for the image-based hand inference task shown in Figure \ref{fig:application_result} (c), we build upon I2L-MeshNet \cite{moon2020i2l} and train another parameter regression branch with 3D joint loss and photometric loss using the FreiHand \cite{zimmermann2019freihand} dataset. 
Note that FreiHand offers ground truth annotation with 21 3D keypoints, while our model is defined with 25 anatomical joints. Following \cite{li2021piano}, we add an additional linear layer that maps from our joint to dataset annotation to account for the mismatch. In addition, we add L2-regularizers on the magnitude of the shape, pose, and appearance parameters.
We assume the fixed lighting condition same as HTML \cite{qian2020html} for predicting appearance parameters. 
The quantitative results are shown in Table \ref{tab:rgb_quan}.
Following \cite{zimmermann2019freihand}, we report PA MPJPE, which is the euclidean distance (mm) between predicted joint and ground truth 3D joint after rigid alignment, as well as F-scores at two different distance thresholds. Though our model does not outperform \cite{moon2020i2l} due to the fundamental difference of joint definition, we are able to achieve a comparable quantitative result and predict unprecedented photorealistic hand with inner structures (Fig.\ref{fig:application_result}(c)).

\begin{table}[]
\caption{Quantitative results of RGB inference. We report joint error with PA MPJPE and F-scores between \cite{moon2020i2l} and our model on FreiHAND dataset.}
\begin{tabular}{c|c|c|c}
\hline
Methods &  PA MPJPE $\downarrow$  & F@5mm $\uparrow$ & F@15mm  $\uparrow$ \\ \hline \hline
\cite{moon2020i2l}   &   \textbf{7.4}   & \textbf{0.681} &  \textbf{0.973}  \\ \hline
\textbf{NIMBLE}    &    9.4   &   0.547  &    0.955       \\ \hline
\end{tabular}
\label{tab:rgb_quan}
\end{table}

\section{Conclusion and Future Work}

To generate faithful hands in Metaverse for immersive experience, we propose a non-rigid hand model, NIMBLE, with skins, bones, and muscle, which is anatomically correct and meets the delicate coordination of inner and outer kinematic structures of hands. Especially, the data we rely on is an enhanced MRI hand dataset with full segmentation annotations for bones, muscles, and skins, as well as auto-registered meshes by our proposed optimization method with physical constraints. For the parameter learning of NIMBLE, we also involve penalty terms to guarantee physically correct muscle deforms. By enforcing inner bones and muscles to match anatomic and kinematic rules, NIMBLE provides an unprecedented level of realism and achieves anatomically correct digital hand synthesis, motion animation and photorealistic rendering. Due to the parametric representations, NIMBLE also benefits many learning-based vision applications with different modalities of input data.

There are several avenues for future work. 
We demonstrate how NIMBLE can change shape and pose with inner and outer consistency, but we do not explicitly model the interconnections. We intend to utilize implicit skinning to include explicit constraints on bone, muscle, and skin interactions.
Besides, with our tetrahedron modeling, we can extend the model to include parametric secondary deformation using specifically designed blend shapes or FEM soft body dynamics, as in \cite{pons2015dyna,tsoli2014breathing}. 
We also want to analyze muscular attributes like stiffness and elasticity to produce a more realistic physical model for efficient muscle and flesh modeling.
We plan to extend our parametric model to include tendons and ligaments to improve skin deformation and overall hand movement realism, allowing for even more dexterous hand modeling and anatomical and kinematics analysis in the future. 
Additionally, we plan to use alternative approaches like geometric modeling via parametric or learning-based methods to model skin wrinkles. 
However, such approaches require capturing significantly more detailed normal maps. Our next step is to utilize the HandStage to capture dynamic sequences to model these fine details on both shape and appearance. 
For hand tracking applications, which highly rely on hand datasets, while existing multiview datasets like \cite{zimmermann2019freihand,inter26} provide limited annotation, and synthetic datasets like \cite{hasson2019learning} lack realism and have domain gap compared to real images.
We plan to utilize NIMBLE to create a high-quality hand dataset with comprehensive ground truth annotation including inner and outer geometry, pose, and appearance, and further train a deep network with it for hand motion capture. 
Finally, two-handed contact and object interaction are also vital. We only use right-handed data, but a left-handed model and hand-object parametric model would be tremendously useful for two-hand motion capture and immersive VR interactions.


\begin{acks}
This work was supported by NSFC programs (61976138, 61977047), the National Key Research and Development Program (2018YFB2100500),  STCSM (2015F0203-000-06), and SHMEC (2019-01-07-00-01-E00003).

\end{acks}

\bibliographystyle{ACM-Reference-Format}
\bibliography{sample-base}

\appendix

\end{document}